%% file: main.tex
\documentclass[manuscript,screen]{acmart}

\setcopyright{none}
\acmJournal{TAAS}
\acmYear{2026}
\acmDOI{}
\acmISBN{}

\usepackage{amsmath}
\usepackage{booktabs}
\usepackage{array}
\usepackage{longtable}
\usepackage{url}
\usepackage{algorithm}
\usepackage{algpseudocode}

\title{Tiny Collaborative Inference for Occlusion-Robust Object Detection}

\author{Chieh-Tung Cheng}
\affiliation{%
  \institution{Imperial College London}
  \city{London}
  \country{United Kingdom}
}

\author{Mustafa Aslanov}
\affiliation{%
  \institution{Nottingham Trent University}
  \city{Nottingham}
  \country{United Kingdom}
}
\email{mustafa.aslanov2025@my.ntu.ac.uk}

\author{Eiman Kanjo}
\affiliation{%
  \institution{Nottingham Trent University}
  \city{Nottingham}
  \country{United Kingdom}
}
\affiliation{%
  \institution{Imperial College London}
  \city{London}
  \country{United Kingdom}
}
\email{eiman.kanjo@ntu.ac.uk}

\begin{document}

\begin{abstract}
Edge AI nodes for search and rescue are increasingly expected to run computer vision locally, yet ultra-low-end hardware imposes hard constraints on memory, compute, and inter-device communication. This work addresses occlusion-robust object detection on devices with less than 1 MB SRAM by combining an MCUNet backbone, a YOLOv2 detection head, and Lite quantisation.
Two collaborative inference strategies are evaluated: feature-level fusion, concatenating intermediate feature maps, and decision-level fusion via Weighted Boxes Fusion (WBF). WBF outperforms feature-level fusion under all tested occlusion conditions, yielding gains of up to +0.2736 mAP in asymmetric scenarios. Extending fusion to three views improves accuracy further (up to +0.3827 mAP) at modest communication overhead (~1.3 KB per exchange).
Hardware experiments progress from a host-assisted USB-relay baseline to a Wi-Fi peer-to-peer deployment on two Coral Dev Board Micro units, where WBF executes on-device with negligible communication energy relative to inference. In a 301.9 s autonomous session of 108 frames, fused output is produced on 61 frames versus 47 for a single board — a coverage gain of +29.8\%. A decentralised federated learning feasibility note is included but not treated as a primary result, as performance remains limited under non-iid data.
The results support decision-level fusion as a viable option for improving occlusion robustness in small-scale edge object detection, including host-free multi-board operation on ultra-low-end hardware.
\end{abstract}



\maketitle

\input{body.tex}
\input{appendices.tex}

\input{refs.tex}
\end{document}

%% file: body.tex
\section{Introduction}

Edge devices such as drones, surveillance cameras, and mobile robots are now common in real-time sensing applications. McKinsey estimates that the economic potential of the Internet of Things (IoT) could reach \$5.5 trillion to \$12.6 trillion globally by 2030 \cite{ref1}. As these devices collect more data at the point of sensing, there is a clear incentive to run AI models locally rather than sending all data to the cloud \cite{ref2}. Edge AI combines the low-latency and privacy benefits of edge computing with the perception capabilities of deep learning. Computer vision (CV) is a natural target for this setting because many edge platforms are built around cameras or other visual sensors.

Running deep learning models on such hardware is still difficult. Ultra-low-end devices usually have no GPU and must operate within tight energy and memory budgets \cite{ref4}. If several devices cooperate, the system also has to manage communication overhead and synchronisation delay under realistic network conditions \cite{ref2}. At the same time, many edge CV applications involve cluttered scenes, changing illumination, and partial occlusion \cite{ref6,ref7,ref8}.

This work studies a lightweight pipeline for occlusion-robust object detection on ultra-low-end edge devices. The design keeps the model small enough for constrained hardware, uses collaborative inference to recover detections that are weak in a single view, and includes a limited decentralised training experiment as an exploratory extension:
\begin{itemize}
\item A lightweight object detection model optimised for ultra-low-resource hardware, such as the Google Coral Dev Board Micro.
\item A collaborative inference setup that compares feature-level and decision-level fusion under controlled occlusion.
\item An exploratory decentralised federated learning (DFL) study for testing whether peer-to-peer adaptation remains numerically stable under non-iid local data.
\end{itemize}

Collaborative inference is the main focus of the paper. Here, it refers to fusing information from multiple edge devices at inference time. DFL is kept separate as a secondary study, where devices exchange model updates during training without a central coordinator.

The paper makes the following contributions:
\begin{itemize}
\item We adapt MCUNet with a YOLOv2 detection head and apply TensorFlow Lite quantisation, enabling efficient object detection on ultra-low-end edge devices.
\item We analyse collaborative inference strategies by comparing feature-level and decision-level fusion under varying occlusion conditions.
\item We extend collaborative inference to multiple views and quantify the trade-off between accuracy gains and the communication overhead introduced by inter-device Wi-Fi transmission.
\item We validate the complete collaborative pipeline on two physical Coral Dev Board Micros, first with a lightweight serial exchange protocol and host-side WBF relay, and then with fully autonomous Wi-Fi peer-to-peer fusion executed on-device.
\item As a secondary exploratory study, we report a preliminary decentralised training experiment based on FedAvg-style peer averaging for lightweight object detection, while explicitly highlighting its current limitations under non-iid data.
\end{itemize}

\section{Literature Review}

\subsection{Tiny Computer Vision Models}\label{subsec:tiny-cv-models}
Liu et al. \cite{ref9} survey several families of lightweight computer vision models, including Shift-based networks, MobileNet, and EfficientNet. These architectures are useful for mobile and embedded systems with moderate resources, but they are still often too large for ultra-low-end microcontrollers (MCUs), where SRAM can be below 1 MB and flash storage is also limited. This mismatch has pushed work toward TinyML and MCU-specific inference libraries such as CMSIS-NN and Tiny Engine, which provide kernels and runtime optimisations designed for small memory footprints. Lin et al. \cite{ref10} address the same constraint with MCUNet, a co-design framework that searches both the network architecture and the inference engine. MCUNet reports 0.707 ImageNet top-1 accuracy and, when combined with a YOLOv2 detection head, improves PASCAL VOC detection accuracy by up to 20\% under a 512 kB SRAM constraint compared with MobileNetV2 baselines. MCUNet V2 \cite{ref11} further reduces peak activation memory through patch-by-patch execution in early CNN layers, allowing larger input resolutions without storing full feature maps. With a YOLOv3 detection head, MCUNet V2 reaches 0.646 mAP under 256 kB SRAM and 0.683 mAP under 512 kB SRAM, a 16.9\% improvement over the previous state of the art. For this paper, the main takeaway from this line of work is that practical MCU vision systems require both compact model architectures and hardware-aware inference execution.

\subsection{Model Compression Techniques}
\begin{figure}[t]
\centering
\includegraphics[width=0.9\columnwidth]{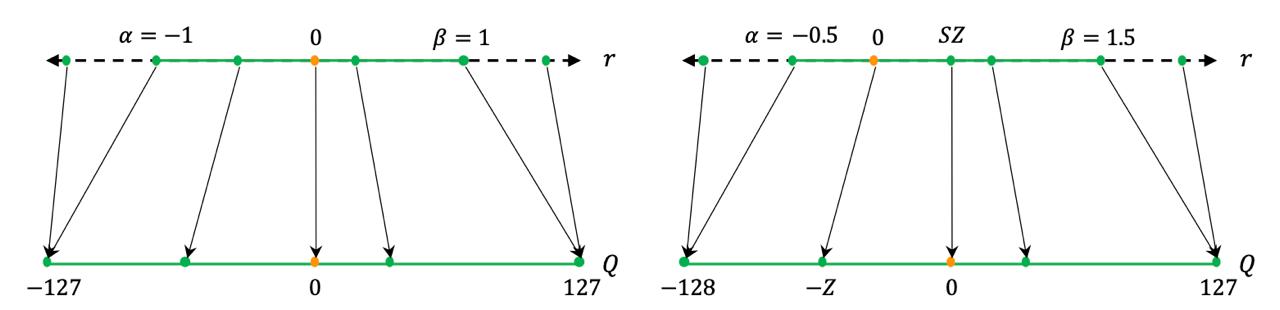}
\caption{Symmetric (left) and asymmetric (right) quantisation representation \cite{ref9}. Note that $r$ represents the real value, $S$ represents the real-valued scaling factor, and $Z$ represents the integer zero point.}
\Description{Two schematic diagrams compare symmetric and asymmetric quantisation. The left diagram shows quantisation centred on zero, and the right diagram shows quantisation with a non-zero zero point. The symbols $r$, $S$, and $Z$ denote the real value, scaling factor, and zero point.}
\label{fig:quantisation-representation}
\end{figure}

 Liu et al. \cite{ref9} group model-compression methods into pruning, quantisation, knowledge distillation, and neural architecture search (NAS). Pruning and distillation can reduce model cost, but they are less central to the hardware setting considered here. Pruning removes redundant weights, yet the remaining parameters are commonly stored in floating-point form unless additional compression is applied. Distillation can improve a small model by training it against a larger teacher, but it introduces teacher selection and extra loss-design choices. Quantisation is more directly relevant because it reduces the precision of weights and activations, usually through calibration, lowering memory use and integer arithmetic cost while aiming to preserve accuracy (Figure~\ref{fig:quantisation-representation}). NAS is also relevant because frameworks such as TinyNAS in MCUNet restrict the search space around MCU-friendly choices, including kernel size, depth, and stride. In this study, quantisation and NAS are therefore treated as the most practical compression tools for the target edge platform.

\subsection{Robust Computer Vision Models under Occlusion}
 Partial occlusion is common in object detection, especially in camera networks where each device has a fixed viewpoint and limited field of view \cite{ref12}. The problem is more difficult for small edge systems because they cannot always compensate by using larger models or higher-resolution inputs. Zhu et al. \cite{ref12} show that CNN accuracy can degrade substantially when objects are severely occluded, even when the same networks perform well on fully visible inputs. Saleh et al. \cite{ref13} review related occlusion-handling strategies across indoor and outdoor detection settings.

\paragraph{Single Model Approach}\mbox{}\\
One response to occlusion is to modify the detector itself. Kortylewski et al. \cite{ref14}, for example, introduce CompositionalNet, which combines deep convolutional features with a generative compositional layer in place of the standard fully connected classification head. This design improves robustness to partial occlusion compared with standard DCNNs.

\paragraph{Single Model Voting Mechanisms}\mbox{}\\
Wang et al. \cite{ref15} extend this idea with a context-aware model that separates object features from contextual features in the image representation. The separation helps reduce the influence of misleading context under occlusion. Their detector also uses part-based voting: individual parts predict object centres and bounding-box offsets, and the part-wise predictions are combined into a final box. Zhang et al. \cite{ref16} use a related idea in DeepVoting, where visible semantic parts learn to vote for the positions of occluded parts. These methods show that internal voting can help a single model retain localisation ability when important object regions are missing.

\paragraph{Ensemble-based Techniques Across Multiple Models}\mbox{}\\
Ensemble methods approach the same problem by combining evidence from several models or views. Balamurugan et al. \cite{ref17} propose a Wrap-CNN ensemble for multi-view object classification, where nine pre-trained CNNs, including ResNet, MobileNetV2, and Xception, process differently rotated views of the same object. A voting scheme then aggregates the model outputs. This type of multi-view voting is relevant to occlusion because features hidden from one viewpoint may remain visible from another. Taken together, the single-model and ensemble studies suggest that combining partial evidence can improve recognition when individual observations are incomplete.
\subsection{Collaborative Inference in CV}
 Collaborative inference extends the ensemble idea to spatially distributed edge devices \cite{ref18}. Instead of relying on one camera or one model output, several devices can cooperate during inference by sharing either intermediate representations or final predictions. In this paper, collaborative inference refers only to inference-time fusion; the base detector is not retrained as part of the fusion step. Following the early/late fusion distinction used in multimodal learning \cite{ref19,ref20}, we consider two categories: feature-level fusion, where views are combined after feature extraction, and decision-level fusion, where independently generated outputs are merged after detection.

\paragraph{Feature-level fusion}\mbox{}\\
Feature-level fusion combines representations before the final prediction layer. In multimodal action recognition, for example, features extracted from RGB images, depth maps, and skeletal sequences can outperform unimodal inputs \cite{ref21}. The cited work also applies a variance-based feature selection step, keeping features whose validation-set variance exceeds a chosen threshold. This reduces the fused representation to more informative features and helps limit overfitting. Although the task differs from object detection, the result motivates testing whether intermediate feature maps from different camera views can provide complementary cues under occlusion.

\paragraph{Decision-level fusion}\mbox{}\\
Decision-level fusion combines final outputs such as logits, class labels, or bounding boxes. Tang et al. \cite{ref22} use a segmentation-based ensemble in which image regions are magnified and processed by parallel YOLO or SSD detectors before the regional predictions are merged. This late-stage merging helps with small or partially visible objects because information from multiple regions is combined after detection. Su et al. \cite{ref23} propose FedOD for cross-domain object detection. Each client maintains a global model and a personalised local model; intermediate features support multi-teacher distillation during training, and Weighted Boxes Fusion (WBF) \cite{ref24} combines global and local predictions during inference. This is useful when client data distributions differ, since the system can use both shared and domain-specific information without full model sharing.

Much of the multi-view literature also considers 3D object recognition. Alzahrani et al. \cite{ref25} review methods that often rely on camera calibration, synchronised capture, and geometric alignment across views. Those assumptions are difficult to satisfy in a small decentralised edge deployment, where cameras may be uncalibrated and communication is limited. For this reason, the present work focuses on 2D detection and lightweight prediction fusion rather than 3D reconstruction. Palena et al. \cite{ref18} have studied collaborative inference for edge devices, but collaborative object detection on ultra-low-end hardware remains less explored, especially under occlusion where different views may contain complementary evidence.
\subsection{Decentralised Federated Learning in CV}
\begin{figure}[t]
\centering
\includegraphics[width=0.9\columnwidth]{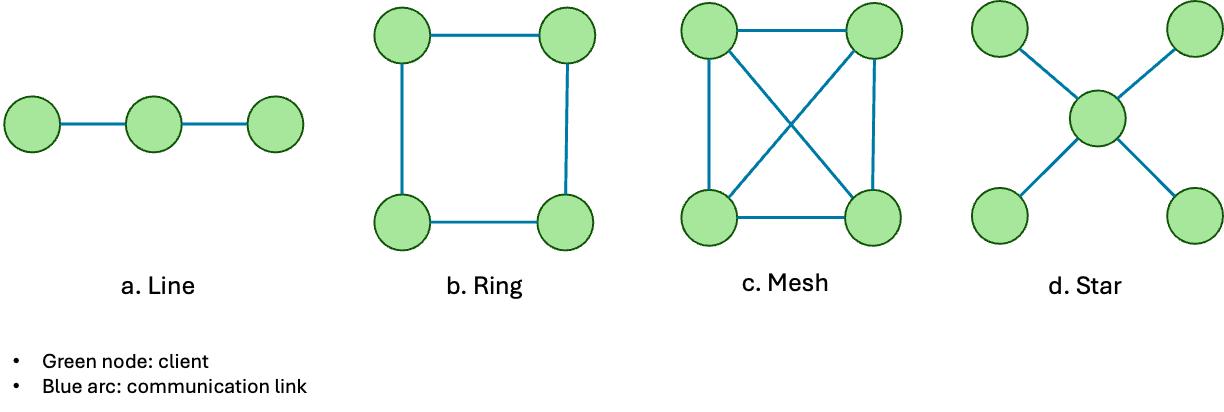}
\caption{Illustration of common network topologies used in DFL}
\Description{Diagram of common decentralised federated learning topologies, showing several ways edge devices can exchange model updates, such as ring-style and more densely connected peer-to-peer layouts.}
\label{fig:dfl-topologies}
\end{figure}

 Decentralised Federated Learning (DFL) removes the central aggregation server used in conventional federated learning. Lalitha et al. \cite{ref26} describe a setting where each client communicates only with neighbours and exchanges model weights or gradients to learn a shared model. Yuan et al. \cite{ref27} organise DFL design around three components: network topology, communication protocol, and learning paradigm. Topology specifies how clients are connected. As shown in Figure~\ref{fig:dfl-topologies}, common choices include Line, Ring, Mesh, and Star configurations. Line and Ring topologies are simple to implement but can introduce sequential bottlenecks and are more sensitive to node failure \cite{ref27}. Mesh topologies can converge faster through denser connectivity, but they also increase communication overhead.

DFL protocols specify how updates move between peers. In pointing communication, a client sends its update to a chosen neighbour \cite{ref27}. Gossip protocols, including the randomized gossip algorithms of Boyd et al. \cite{ref28}, let nodes exchange information with randomly selected peers over time so that parameters diffuse through the network. Yuan et al. \cite{ref27} also distinguish between Aggregate and Continual learning paradigms. In the Aggregate paradigm, a client collects several peer updates, averages or otherwise aggregates them, and then continues local training. In the Continual paradigm, a model is passed from one peer to the next, with each client updating it in sequence. Aggregate methods are closer to classical federated learning but use more communication, whereas Continual methods reduce communication but can be sensitive to update order and forgetting.

For edge computer vision, DFL is attractive mainly because it avoids assuming a stable high-bandwidth connection to a central server. Himeur et al. \cite{ref29} review federated learning in computer vision for devices such as UAVs, smartphones, and surveillance cameras, but much of the work still focuses on centralised federated learning (CFL). That assumption can be weak in infrastructure-limited or mobile settings such as SAR operations, where devices may disconnect or encounter changing visual conditions. In this paper, DFL is therefore considered as a possible adaptation mechanism for collaborative edge systems, but only as a small feasibility study rather than as a mature deployment result.

\section{Problem Description}

The review above leaves three issues that are directly relevant to this study. First, tiny vision models and compression frameworks such as MCUNet show that deployment below 1 MB SRAM is possible, but their behaviour under partial occlusion is less well characterised. Second, collaborative inference can help when a target is partly hidden, but most existing studies do not test this idea on ultra-low-end devices running tiny detectors. Third, field deployments may require adaptation to changing local conditions, while DFL for lightweight object detection remains much less developed than CFL. This paper therefore evaluates MCUNet with quantisation, compares feature-level and decision-level fusion under controlled occlusion, measures the accuracy--communication trade-off when adding views over Wi-Fi, validates the pipeline on two Coral Dev Board Micros in both USB-relay and Wi-Fi peer-to-peer modes, and reports a small FedAvg-style DFL experiment under non-iid edge data.

\section{System Design}

\subsection{System Architecture Overview.}
\begin{figure}[t]
\centering
\includegraphics[width=0.9\columnwidth]{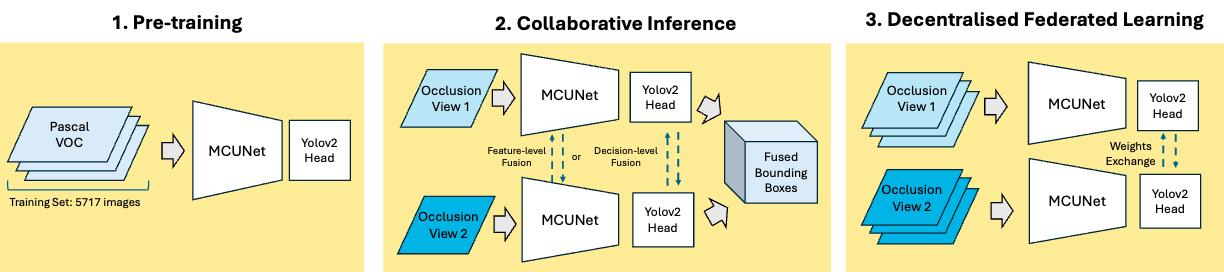}
\caption{Overview of the system architecture: (Left) model pre-training with MCUNet backbone and YOLOv2 head; (Middle) collaborative inference using feature-level or decision-level fusion of occluded views; (Right) DFL with weight exchange across devices}
\Description{Three-part system diagram. The left panel shows pre-training with an MCUNet backbone and YOLOv2 head. The middle panel shows collaborative inference from multiple occluded views using feature-level or decision-level fusion. The right panel shows decentralised federated learning with model-weight exchange between devices.}
\label{fig:system-overview}
\end{figure}

 As shown in Figure~\ref{fig:system-overview}, the proposed system architecture consists of three main stages: pre-training, collaborative inference, and decentralised federated learning. In the pre-training stage, a lightweight object detection model is fine-tuned on a general dataset of primarily non-occluded objects. This step equips the model with basic detection capability across diverse and generic scenarios. Once deployed in the field, the collaborative inference stage begins. Each edge device performs inference on its locally captured images, which may include partially occluded views. For each image pair observed by the devices, a fusion strategy, either feature-level fusion or decision-level fusion, is applied to combine feature representations or predictions into a unified set of detections. Finally, after several rounds of collaborative inference, the system enters the decentralised federated learning phase. In this stage, each device uses its locally captured occluded images together with ground-truth annotations to perform supervised local fine-tuning. To enable distributed adaptation, we adopt FedAvg-style peer averaging within the pointing protocol and ring topology, where devices periodically exchange model weights with their neighbours.

\subsection{Pre-training Strategy}\label{subsec:system-pretraining}
\begin{figure}[t]
\centering
\includegraphics[width=0.9\columnwidth]{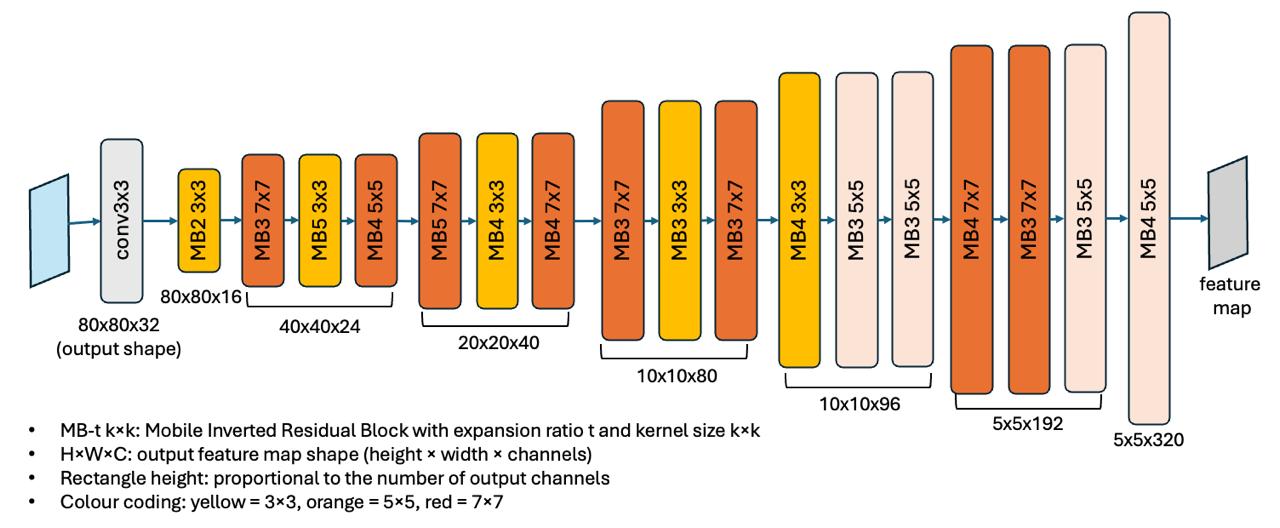}
\caption{Backbone structure of MCUNet}
\Description{Block diagram of the MCUNet backbone, showing a sequence of lightweight convolutional stages that reduce spatial resolution while increasing channel depth.}
\label{fig:mcunet-backbone}
\end{figure}

\paragraph{Model Selection}\mbox{}\\
 As discussed in Section~\ref{subsec:tiny-cv-models}, MCUNet V2 provides a highly efficient architecture designed for resource-constrained devices. In this work, however, we adopt MCUNet as the backbone due to the availability of open-source pre-trained weights and stronger community support, which make it more practical for deployment and reproducibility. This choice ensures that our framework can be easily replicated and extended by others. The overall backbone structure of MCUNet is illustrated in Figure~\ref{fig:mcunet-backbone}.

\begin{figure}[t]
\centering
\includegraphics[width=0.9\columnwidth]{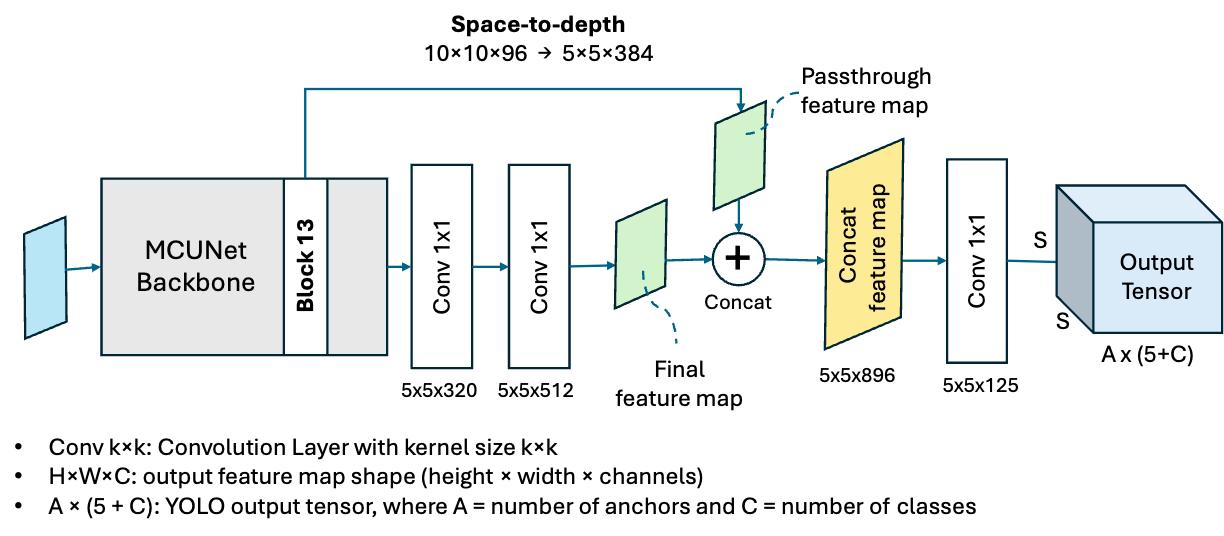}
\caption{YOLOv2 detection head used in this work}
\Description{Block diagram of the YOLOv2 detection head used in the study, including passthrough features, feature concatenation, intermediate convolution layers, and the final prediction output.}
\label{fig:yolov2-head}
\end{figure}

We adopt the YOLOv2 \cite{ref30} detection head, which is both lightweight and memory efficient compared to later YOLO variants. A key feature of YOLOv2 is the passthrough layer, designed to enhance small object detection. Specifically, we extract a higher-resolution $10 \times 10 \times 96$ feature map from the 13th block of MCUNet and apply a space-to-depth operation, which rearranges it into a $5 \times 5 \times 384$ tensor. This tensor is then concatenated with the final $5 \times 5 \times 320$ feature map produced by the second convolutional layer in the detection head. The combined representation compensates for the coarse resolution of the final layer, which is effective for large objects but insufficient for detecting small ones. As illustrated in Figure~\ref{fig:yolov2-head}, the detection head consists of a 320-channel convolutional layer followed by a 512-channel convolutional layer. During the VOC pre-training stage, the concatenated feature map is processed by a 125-channel output layer, corresponding to $A \times (5 + C) = 5 \times (5 + 20)$ for five anchors and 20 VOC classes. For the later car-only collaborative inference and hardware deployment stages, this output head is reconfigured to 30 channels, i.e. $5 \times (5 + 1)$, to match the single-class setting. These outputs encode the bounding box regression, objectness score, and class scores for each anchor. The loss function adopts the original YOLOv2 formulation, consisting of three components: coordinate regression, objectness confidence, and classification. The formal loss function is detailed in Appendix~\ref{app:yolov2-loss}. 
\paragraph{Dataset}\mbox{}\\ 
The PASCAL VOC is one of the most widely used benchmarks for object detection, containing 20 object categories with 5,717 images in the training set and 5,823 images in the validation set. It has frequently been adopted in edge-device object detection research \cite{ref31}. In this work, VOC is used to enhance the generalisation capability of the pre-trained model. By training on VOC, the object detector can first develop a robust understanding of objects under relatively controlled conditions, such as limited occlusion and clearer appearances, thus providing a foundation for subsequent evaluation in more complex field scenarios.

\paragraph{Evaluation}\mbox{}\\
 We evaluate the MCUNet backbone against two widely used lightweight architectures: MobileNetV2, which is optimised for efficiency on resource-constrained devices, and ResNet-18, a small residual network commonly adopted as a baseline in computer vision research. This comparison allows us to assess whether MCUNet is the most suitable backbone when integrated with a YOLOv2 detection head. Performance is measured using mean Average Precision at an Intersection over Union threshold of 0.5 (mAP@0.5). This setup demonstrates the trade-off between detection accuracy and model complexity. 
 \subsection{Collaborative Inference.}\label{subsec:system-collab}
\begin{figure}[t]
\centering
\includegraphics[width=0.9\columnwidth]{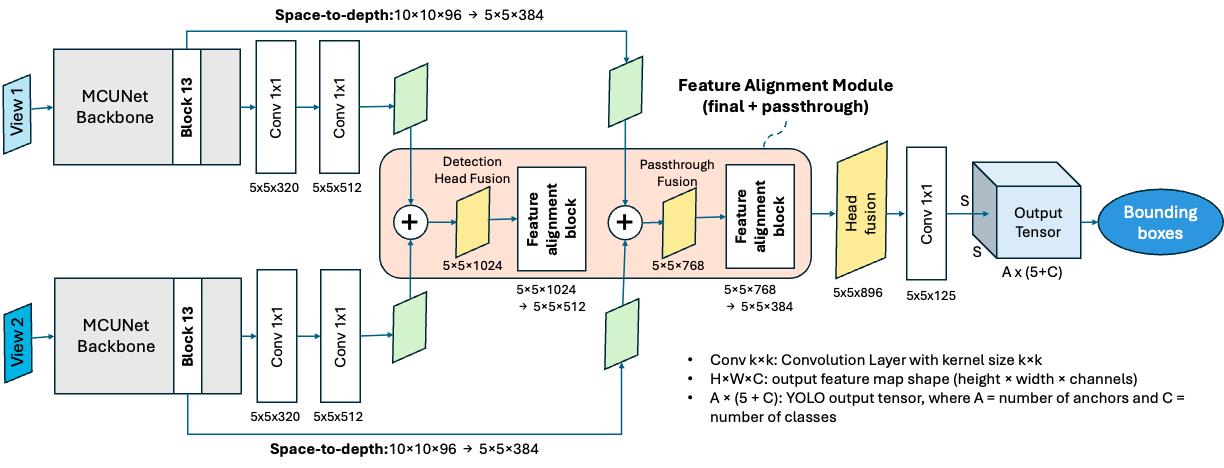}
\caption{Multi-view feature-level fusion pipeline}
\Description{Pipeline diagram for multi-view feature-level fusion. Two camera views are processed separately, intermediate feature maps are exchanged and concatenated, and a joint detector produces the final prediction.}
\label{fig:feature-fusion-pipeline}
\end{figure}

\begin{figure}[t]
\centering
\includegraphics[width=0.9\columnwidth]{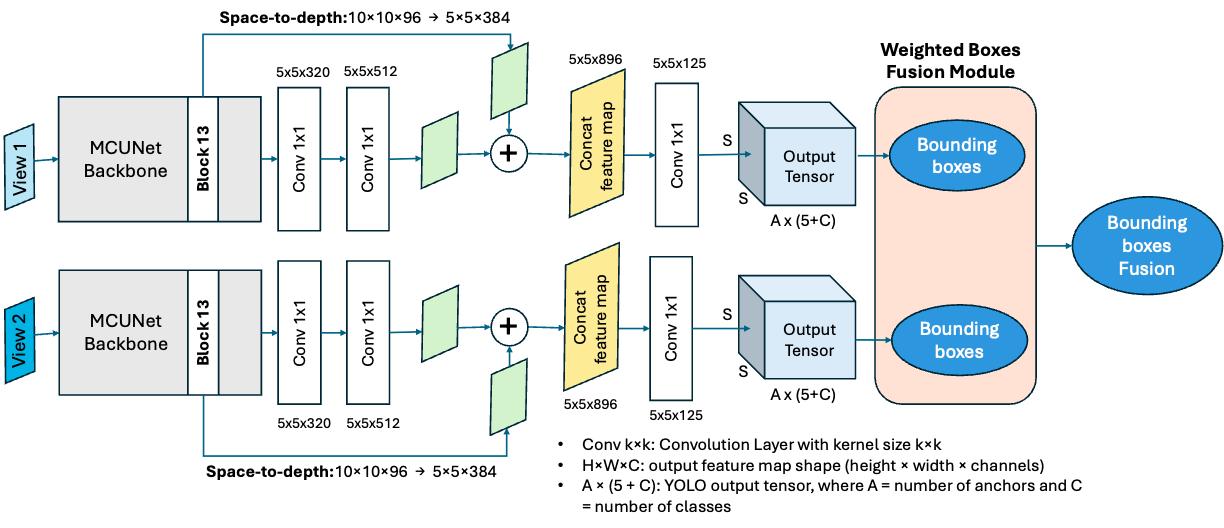}
\caption{Multi-view decision-level fusion pipeline}
\Description{Pipeline diagram for multi-view decision-level fusion. Each view is processed independently to produce detections, and the final bounding boxes are merged at the output stage using Weighted Boxes Fusion.}
\label{fig:decision-fusion-pipeline}
\end{figure}

  Due to the computational and communication constraints of edge devices, our system adopts two practical collaborative strategies that avoid exchanging raw inputs: feature-level fusion and decision-level fusion. These approaches aim to leverage complementary information across multiple viewpoints to improve detection performance under occlusion. The overall pipelines for both strategies are shown in Figures~\ref{fig:feature-fusion-pipeline} and~\ref{fig:decision-fusion-pipeline}, respectively. The fusion stages are highlighted with pink module blocks, indicating where cross-view information exchange occurs in the pipeline.

\paragraph{Feature-Level Fusion: Feature map concatenation}\mbox{}\\
 In feature-level fusion, devices exchange intermediate feature maps from the passthrough layer and detection head, which are combined into a joint representation before the final prediction. To facilitate effective integration, a feature alignment block, which contains a convolutional layer and a batch normalisation layer, is added after fusion to align information from different viewpoints. This allows complementary cues from different viewpoints to be integrated before prediction, enabling the network to learn more robust representations under occlusion. 
 \paragraph{Decision-Level Fusion: Weighted Fusion Boxes}\mbox{}\\ At the decision level, we adopt the Weighted Boxes Fusion (WBF) algorithm \cite{ref24} to merge output predictions (i.e., bounding boxes and confidence scores) from multiple views of the same object. The algorithm clusters bounding boxes whose Intersection over Union (IoU) exceeds a predefined threshold, indicating sufficient overlap. Once all clusters are formed, the final fused bounding boxes are computed by taking the confidence-weighted average of the coordinates and the mean confidence score within each cluster. The step-by-step procedure is presented as Algorithm~\ref{alg:wbf} in Appendix~\ref{app:wbf}. This post-prediction fusion strategy enables the integration of complementary detections from different viewpoints, thereby enhancing robustness without modifying the feature extraction or detection pipeline. 
 \paragraph{Dataset}\mbox{}\\
  The goal of our experiment is to evaluate collaborative inference under controlled occlusion for object detection rather than broad multi-class recognition. We therefore use the CO3D dataset \cite{ref32} but restrict evaluation to a smaller subset of the car category, which provides a rich and consistent semantic structure while keeping the experimental footprint manageable; the full CO3D car category is already approximately 24 GB. To create the occlusion setup, we sample 10 car instances from CO3D and select 10 views per instance. Each view is captured at approximately 35-degree intervals around the object. For each instance, view pairs are formed with small angular differences to ensure overlapping fields of view and minimise multi-view calibration issues. To simulate partial visibility, we apply the Cut Out augmentation technique \cite{ref33}, which masks random regions of each image. This provides artificial control over occlusion levels and enables controlled experiments for collaborative detection. 
  \paragraph{Evaluation}\mbox{}\\
   We compare single-view detection with two multi-view collaborative inference approaches: feature-level fusion and decision-level fusion. The evaluation is conducted under various combinations of occlusion levels. By testing across these scenarios, we aim to determine which fusion strategy is better suited to different levels of occlusion. Model performance is measured using the standard object detection metric, mAP@0.5. 
   \subsection{Decentralised Federated Learning.}
   To explore whether decentralised training remains feasible under dynamic visual conditions, we include a preliminary DFL study in which models adapt incrementally through peer-to-peer weight exchange. This allows each device to learn from heterogeneous local views (e.g., varying occlusion levels) while keeping the setup small and controlled. In this stage, we consider a system of two edge devices connected via a ring topology and communicating through a pointing protocol in each update cycle. This design is suitable for small-scale deployments with predictable communication patterns. For optimisation, we adopt FedAvg-style peer averaging \cite{ref34}, in which model parameters are updated synchronously in a peer-to-peer manner without reliance on a central server. 
   \paragraph{Dataset}\mbox{}\\ To evaluate the convergence behaviour of our decentralised training protocol, we reuse the CO3D subset employed in our inference evaluation. Each image is paired with its corresponding ground-truth annotation, with a total of 171 car instances. Given the limited dataset scale, this experiment is positioned as a feasibility study rather than a benchmarking exercise. 
   \paragraph{Evaluation}\mbox{}\\
    We analyse the convergence trend of the DFL training process to assess whether decentralised averaging remains numerically stable for lightweight object detection in a small non-iid setting. Given the limited dataset scale, this stage is treated as a feasibility study rather than a full benchmark, so we focus on training-loss dynamics as the primary indicator of behaviour.

\section{Experiment Results and Discussion}

\subsection{Pre-training Strategy}\label{subsec:results-pretraining}

\paragraph{MCUNet-YOLOv2 Model}\mbox{}\\
We fine-tune MCUNet-YOLOv2 with the hyperparameters listed in Appendix~\ref{app:mcunet-hparams}. As shown in Figure~\ref{fig:loss-curve}, training converges stably within 160 epochs. After a rapid decrease during the first 10 epochs, the loss continues to decline gradually until plateauing around epoch 120. This indicates that the chosen optimiser and configuration achieve smooth convergence without oscillation or divergence. The stable loss trend suggests that the model is sufficiently optimised and provides a reliable backbone for subsequent experiments.

\begin{figure}[t]
\centering
\includegraphics[width=0.9\columnwidth]{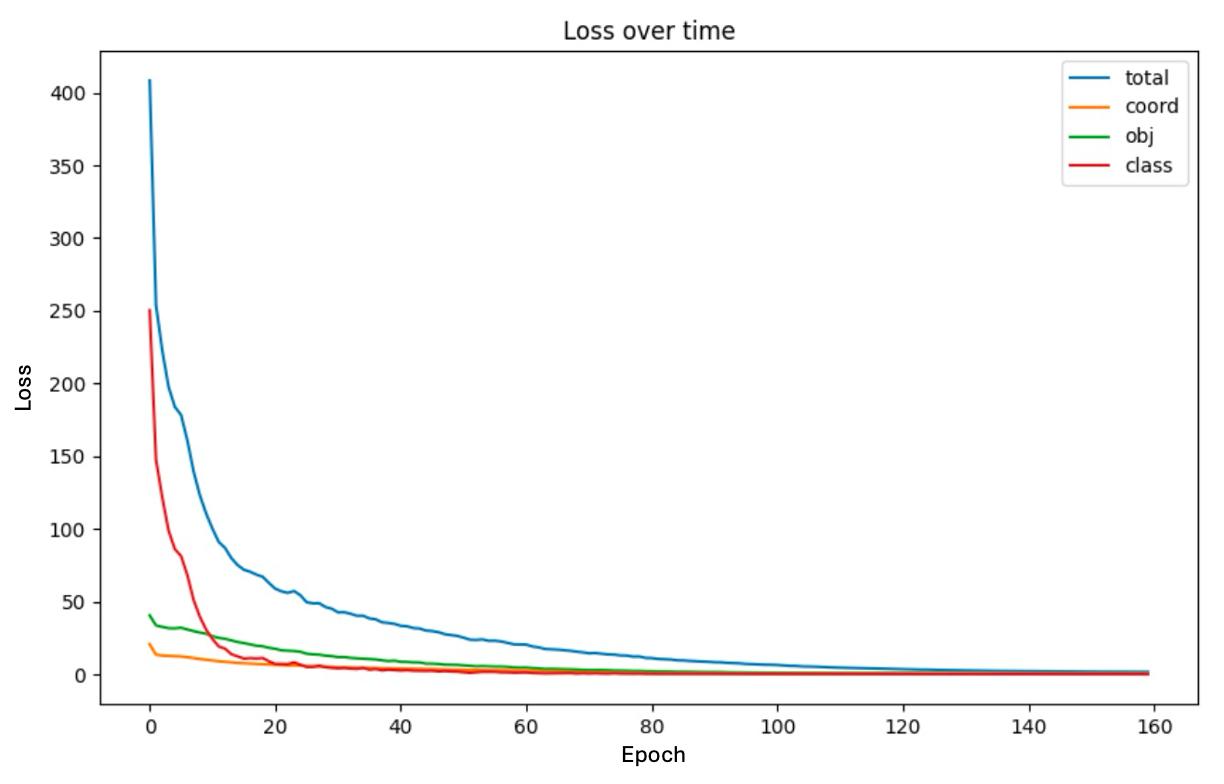}
\caption{Loss curve for fine-tuning MCUNet-YOLOv2 with $160 \times 160$ resolution input.}
\Description{Line plot of training loss against epoch for MCUNet-YOLOv2 fine-tuning at 160 by 160 input resolution, showing a steep initial decrease followed by gradual convergence.}
\label{fig:loss-curve}
\end{figure}

Table~\ref{tab:input-resolutions} summarises the effect of varying the input resolution for MCUNet-YOLOv2. As expected, higher resolutions generally improve detection accuracy, with mAP increasing from 0.2055 at $128 \times 128$ to 0.3096 at $256 \times 256$. However, these gains come at the cost of greater computational complexity and memory usage. FLOPs increase nearly fourfold, from 205.41M to 821.64M, while peak RAM usage rises from 39.7 MB to 45.5 MB across this range.

\begin{table}[t]
\centering
\caption{Comparison of input resolutions for MCUNet-YOLOv2.}
\label{tab:input-resolutions}
\begin{tabular}{@{}ccccc@{}}
\toprule
Resolution & Grid size & mAP@0.5 & FLOPs (M) & Peak RAM (MB) \\
\midrule
$128 \times 128$ & 4 & 0.2055 & 205.41 & 39.7 \\
$160 \times 160$ & 5 & 0.2575 & 320.95 & 40.8 \\
$192 \times 192$ & 6 & 0.2780 & 462.17 & 42.1 \\
$224 \times 224$ & 7 & 0.3052 & 629.07 & 43.7 \\
$256 \times 256$ & 8 & 0.3096 & 821.64 & 45.5 \\
\bottomrule
\end{tabular}
\\[2pt]
\footnotesize{Measured on GPU during inference.}
\end{table}

\begin{figure}[t]
\centering
\includegraphics[width=0.9\columnwidth]{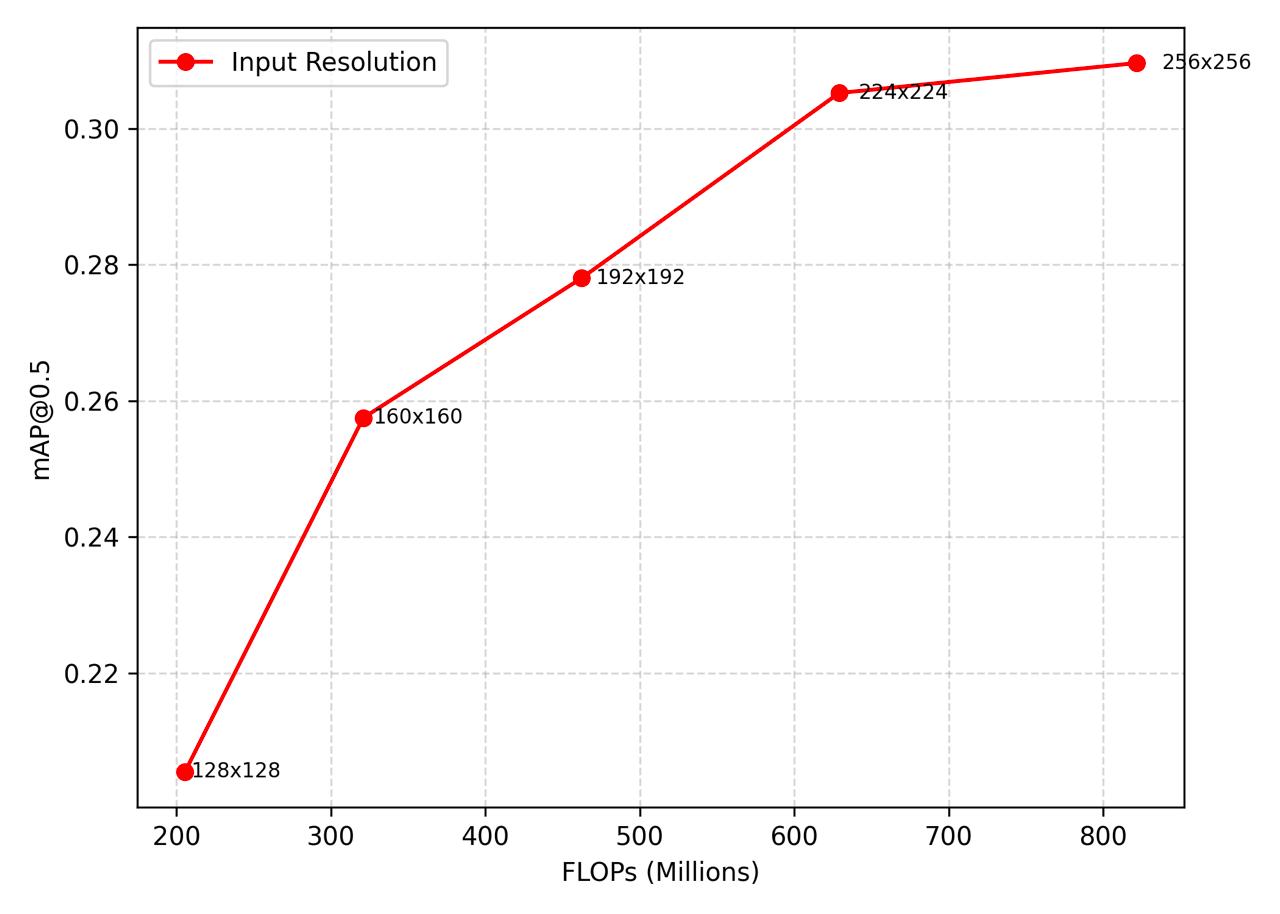}
\caption{Trade-off between accuracy and FLOPs under varying input resolutions.}
\Description{Plot comparing detection accuracy and computational cost in FLOPs across several input resolutions, showing that higher resolution improves accuracy but increases compute substantially.}
\label{fig:resolution-flops-tradeoff}
\end{figure}

As shown in Figure~\ref{fig:resolution-flops-tradeoff}, increasing the resolution from $160 \times 160$ to $192 \times 192$ raises FLOPs by about $1.4\times$, yet yields only a marginal gain of 0.02 mAP. Considering this trade-off, we adopt $160 \times 160$ as the standard resolution for subsequent experiments, as it provides a balanced compromise between accuracy and efficiency.

\begin{figure}[t]
\centering
\includegraphics[width=0.9\columnwidth]{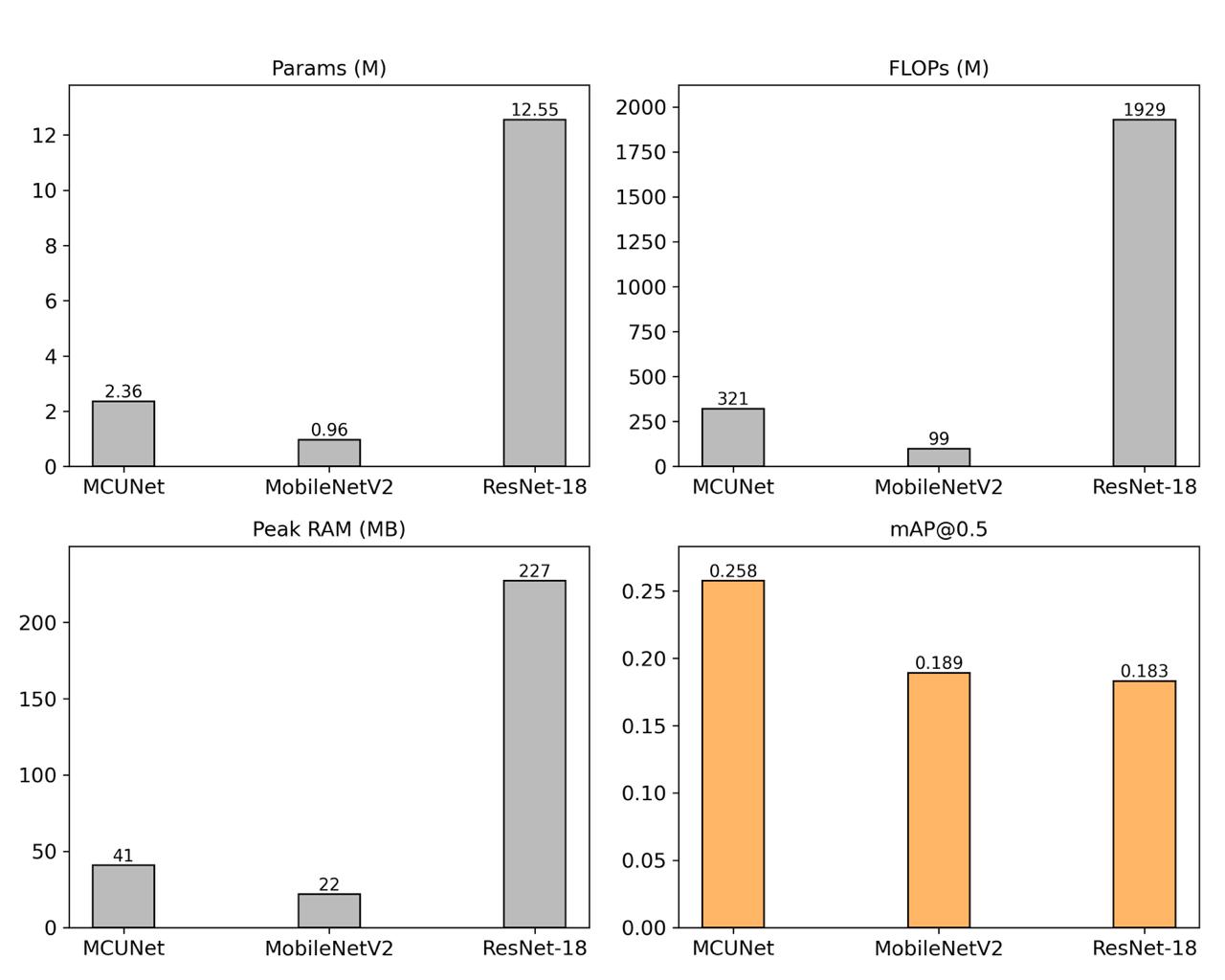}
\caption{Backbone comparison: MCUNet, MobileNetV2, and ResNet-18.}
\Description{Comparison chart for MCUNet, MobileNetV2, and ResNet-18, summarising their relative detection accuracy and efficiency and highlighting MCUNet as the strongest trade-off.}
\label{fig:backbone-comparison}
\end{figure}

We further compare different backbone architectures under the same setting to assess whether MCUNet offers advantages over other widely used lightweight backbones, such as MobileNetV2 and ResNet-18. The comparison results are presented in Figure~\ref{fig:backbone-comparison}. Among the three backbones, ResNet-18 has the largest parameter count and computational cost, yet its detection accuracy at $160 \times 160$ input resolution is slightly lower than that of MobileNetV2. This suggests that ResNet-18 is better suited for higher-resolution inputs and is ineffective in low-resolution settings. MobileNetV2, on the other hand, achieves the smallest model size, lowest FLOPs, and lowest peak memory usage, but its accuracy lags behind MCUNet. MCUNet provides a stronger balance between accuracy and efficiency, delivering the best detection performance while maintaining moderate model size and memory usage. Considering this trade-off, we select MCUNet as the primary backbone for subsequent quantisation experiments.

\paragraph{Quantisation}\mbox{}\\
Although MCUNet proved to be the most suitable backbone among its alternatives, the MCUNet-YOLOv2 model is still too large for direct deployment on MCUs. In particular, the full-precision model exceeds the strict memory and compute constraints of devices with less than 1 MB SRAM. To address these limitations, we apply post-training quantisation, which compresses both weights and activations into lower-precision formats while aiming to preserve detection accuracy.

Following the workflow of~\cite{ref10,ref31}, we convert the pre-trained PyTorch weights into TensorFlow format and quantise the model from 32-bit floating point (FP32) to 8-bit integer (INT8) using the TensorFlow Lite converter. Quantisation maps continuous values to a discrete integer range via a scale factor and zero-point offset, enabling efficient inference on MCUs that support only integer arithmetic.

As shown in Table~\ref{tab:quantised-vs-original}, quantisation results in only a negligible accuracy drop of 0.003 mAP@0.5 on the VOC validation set, while reducing model size and peak RAM usage by approximately 71\% and 83\%, respectively. These improvements demonstrate that the quantised model is feasible for deployment on ultra-low-end MCUs under real-world constraints.

\begin{table}[t]
\centering
\caption{Comparison of quantised and original MCUNet-YOLOv2.}
\label{tab:quantised-vs-original}
\begin{tabular}{@{}lccc@{}}
\toprule
Quantisation scheme & mAP@0.5 & Model storage (MB) & Peak RAM (MB) \\
\midrule
FP32 & 0.2575 & 9.01 & 15.27 \\
INT8 & 0.2545 & 2.61 & 2.55 \\
\bottomrule
\end{tabular}
\\[2pt]
\footnotesize{Measured on CPU during inference.}
\end{table}

\paragraph{Microcontroller Deployment}\mbox{}\\
For on-device evaluation, Arm Cortex-M series CPUs are widely adopted in TinyML research as representative deployment endpoints~\cite{ref10,ref11,ref31}. Accordingly, we select the Google Coral Dev Board Micro as our evaluation platform~\cite{ref37}. It integrates an NXP i.MX RT1176 with Cortex-M7 (800 MHz) and Cortex-M4 (400 MHz) cores, and provides a C++-based development ecosystem (Coral Micro SDK) that facilitates implementation. With 2 MB of built-in SRAM, the Coral Dev Board Micro offers a practical test platform for assessing the feasibility of TinyML models under real deployment conditions.

The results in Tables~\ref{tab:on-device-memory} and~\ref{tab:on-device-latency} show that the model ran reliably in our tests within the 1 MB SRAM constraint. For context, the original MCUNet work reports inference time of around 1.1 s on a Cortex-M7 processor using their co-designed TinyNAS + TinyEngine pipeline~\cite{ref10}. In our case, the observed latency is higher (around 3.2 s), which can be attributed to the use of a YOLOv2 detection head and reliance on a standard TFLite post-training quantisation pipeline rather than a quantisation-aware training framework. Nevertheless, the results demonstrate the feasibility of deploying MCUNet-family models on real-world MCU platforms, offering a practical pathway with minimal engineering overhead.

\begin{table}[t]
\centering
\caption{On-device inference memory usage on the Coral Dev Board Micro.}
\label{tab:on-device-memory}
\begin{tabular}{@{}cccc@{}}
\toprule
Arena used (KB) & Input tensor (KB) & Output tensor (KB) & Intermediate tensors (KB) \\
\midrule
759.2 & 76.8 & 3.125 & 697.5 \\
\bottomrule
\end{tabular}
\end{table}

\begin{table}[t]
\centering
\caption{On-device inference latency on the Coral Dev Board Micro.}
\label{tab:on-device-latency}
\begin{tabular}{@{}cccc@{}}
\toprule
Avg latency (ms) & Min latency (ms) & Max latency (ms) & Invoke errors \\
\midrule
3197 & 3196 & 3200 & 0 \\
\bottomrule
\end{tabular}
\end{table}

\subsection{Collaborative Inference}\label{subsec:results-collab}

\paragraph{Two-view Fusion Results}\mbox{}\\
To evaluate the effectiveness of collaborative fusion, we begin with the two-device setting. We compare the two strategies introduced in Section~\ref{subsec:system-collab}: decision-level fusion using WBF and feature-level fusion by concatenating intermediate feature maps. Experiments are conducted under three occlusion configurations: (30\%, 30\%), (30\%, 50\%), and (50\%, 50\%), where the percentages indicate the level of occlusion in View~1 and View~2. Detection performance is reported using the mAP@0.5 metric.

Across all settings, decision-level fusion consistently outperforms the single-view baseline, which confirms its robustness under varying degrees of occlusion. By contrast, feature-level fusion achieves comparable gains to decision-level fusion in the mixed-occlusion case (30\%, 50\%), but leads to accuracy degradation in the symmetric occlusion scenarios (30\%, 30\%) and (50\%, 50\%). As shown in Table~\ref{tab:two-view-occlusion}, WBF delivers improvements of up to +0.2736 mAP under the (30\%, 50\%) configuration, while maintaining positive gains in all cases. Feature-level fusion can achieve improvements of +0.2625 mAP in the same setting, but exhibits greater instability overall, with drops of up to --0.1769 mAP in the (30\%, 30\%) setting.

\begin{table}[t]
\centering
\caption{Detection performance under different occlusion settings.}
\label{tab:two-view-occlusion}
\resizebox{\columnwidth}{!}{%
\begin{tabular}{@{}llccccc@{}}
\toprule
Occ. pair & View & Baseline & Decision fusion & $\Delta$ (WBF) & Feature fusion & $\Delta$ (Feat.) \\
\midrule
30\%, 30\% & View 1 & 0.5845 & 0.7393 & +0.1548 & 0.5696 & -0.0149 \\
30\%, 30\% & View 2 & 0.6612 & 0.7232 & +0.0620 & 0.4843 & -0.1769 \\
\midrule
30\%, 50\% & View 1 & 0.5845 & 0.6582 & +0.0737 & 0.5942 & +0.0097 \\
30\%, 50\% & View 2 & 0.2758 & 0.5494 & +0.2736 & 0.5383 & +0.2625 \\
\midrule
50\%, 50\% & View 1 & 0.2877 & 0.3411 & +0.0534 & 0.2026 & -0.0851 \\
50\%, 50\% & View 2 & 0.2758 & 0.3747 & +0.0989 & 0.1727 & -0.1031 \\
\bottomrule
\end{tabular}%
}
\\[2pt]
\footnotesize{All values report mAP@0.5. $\Delta$ indicates the performance change relative to the baseline.}
\end{table}

\begin{figure}[t]
\centering
\includegraphics[width=0.9\columnwidth]{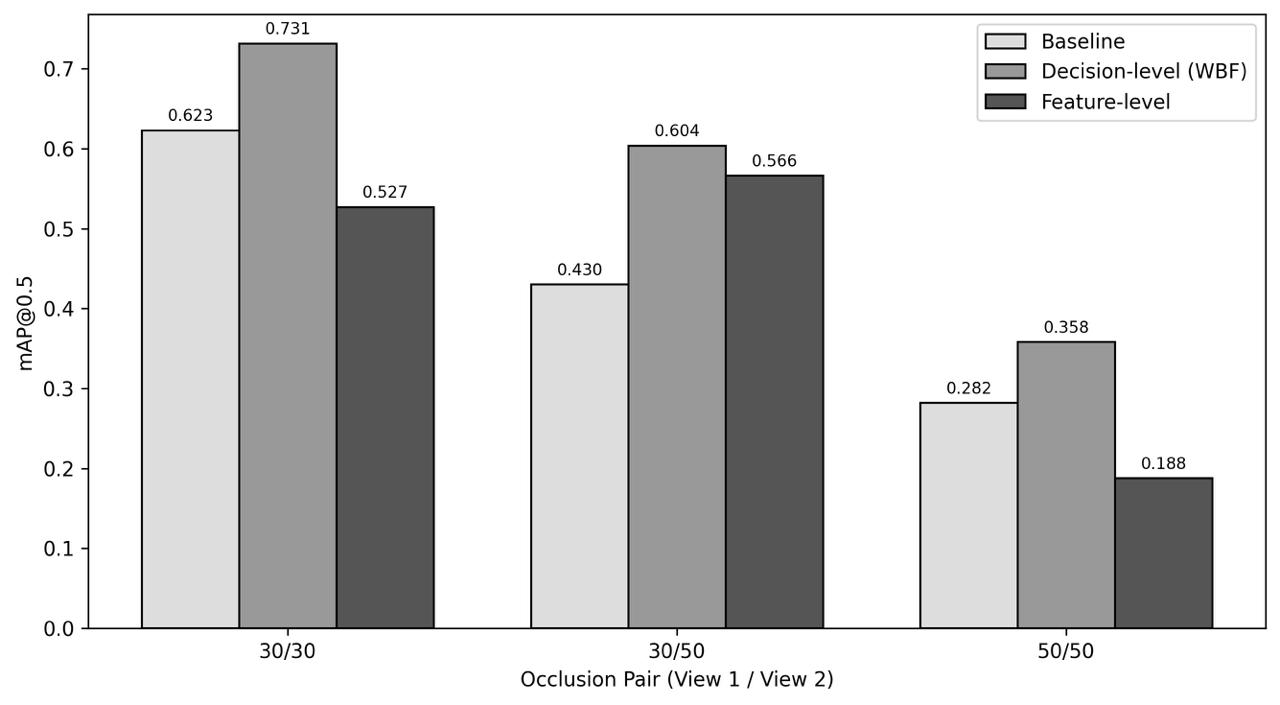}
\caption{Feature-level vs. decision-level fusion accuracy across occlusion pairs.}
\Description{Comparison chart of baseline, feature-level fusion, and decision-level fusion accuracy across several two-view occlusion pairs, showing decision-level fusion as the most consistently effective method.}
\label{fig:two-view-fusion-comparison}
\end{figure}

In Figure~\ref{fig:two-view-fusion-comparison}, we plot the average mAP of single-view, decision-level fusion, and feature-level fusion under three occlusion configurations. The bar chart highlights the overall trend more clearly: decision-level fusion consistently outperforms both the single-view baseline and feature-level fusion across all settings. Notably, the gains of both fusion methods are most pronounced under asymmetric occlusion, demonstrating their ability to leverage information from the less occluded view to compensate for the more occluded one.

We also observe that feature-level fusion can underperform compared with the single-view baseline. A likely explanation is the lack of sophisticated alignment mechanisms in our current implementation. The fusion relies solely on batch normalisation to merge feature maps, rather than more advanced techniques such as feature calibration~\cite{ref35}. These limitations suggest that the degradation may result from the simplicity of the merging strategy. With improved alignment and preprocessing, feature-level fusion could emerge as a competitive alternative for occlusion-robust inference in future work.

\begin{figure}[t]
\centering
\includegraphics[width=0.9\columnwidth]{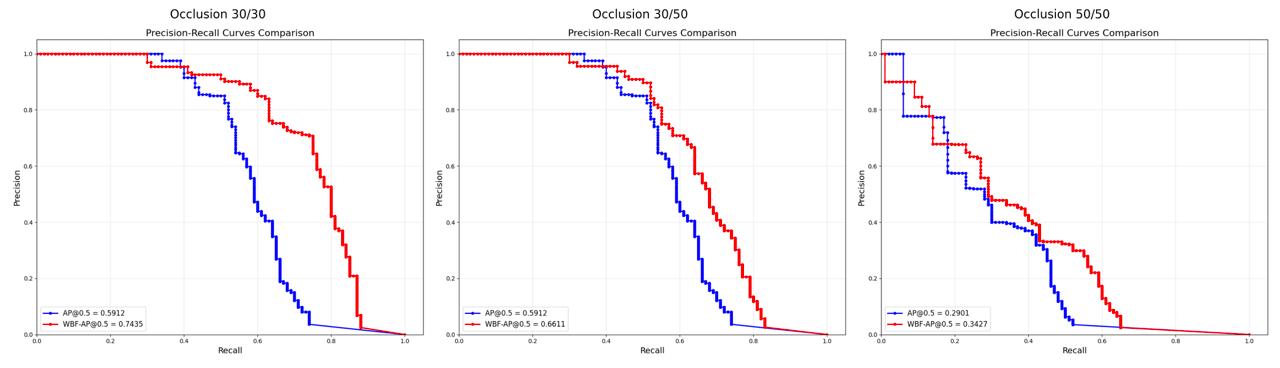}
\caption{WBF (red) and baseline (blue) precision--recall comparison (view 1).}
\Description{Precision--recall curves for View 1 comparing the baseline detector with Weighted Boxes Fusion, with the WBF curve lying above the baseline for most of the operating range.}
\label{fig:pr-view1}
\end{figure}

\begin{figure}[t]
\centering
\includegraphics[width=0.9\columnwidth]{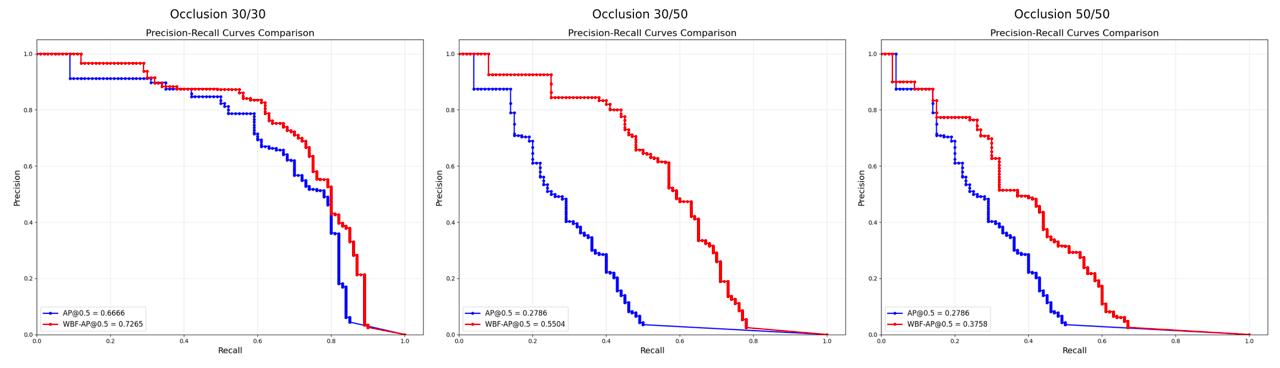}
\caption{WBF (red) and baseline (blue) precision--recall comparison (view 2).}
\Description{Precision--recall curves for View 2 comparing the baseline detector with Weighted Boxes Fusion, again showing the WBF curve above the baseline across most thresholds.}
\label{fig:pr-view2}
\end{figure}

\begin{figure}[t]
\centering
\includegraphics[width=0.9\columnwidth]{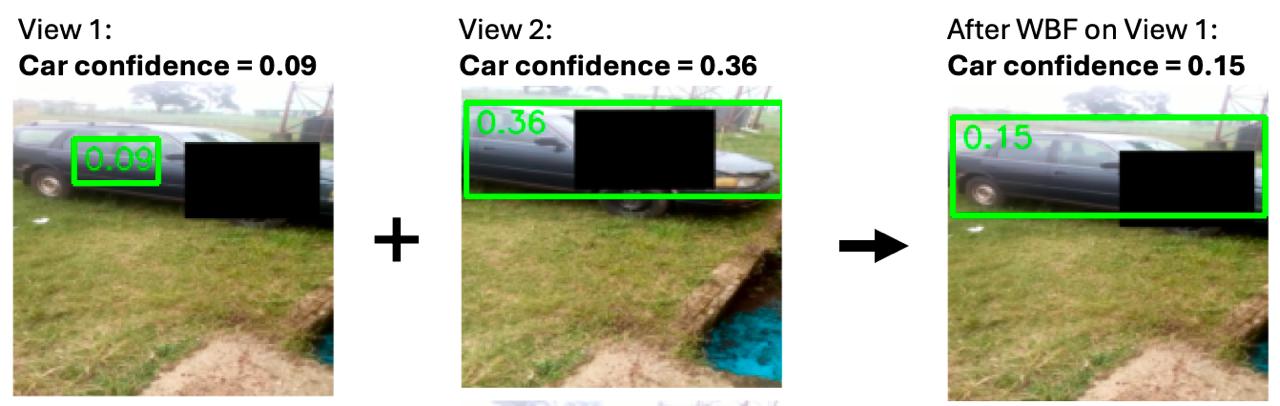}
\caption{Example of confidence averaging in WBF.}
\Description{Illustration of two overlapping detections being merged by Weighted Boxes Fusion into a single consensus box with averaged coordinates and confidence score.}
\label{fig:wbf-confidence}
\end{figure}

We further investigate how complementary views compensate for weaker detections. The precision--recall curves (Figures~\ref{fig:pr-view1} and~\ref{fig:pr-view2}) reinforce our earlier observation that fusion gains are most pronounced under asymmetric occlusion, such as the (30\%, 50\%) setting for View~2. Furthermore, in the right-hand portion of the curves, which corresponds to lower confidence thresholds, WBF consistently achieves higher recall than the single-view baseline. This behaviour demonstrates that fusion increases coverage by aggregating detections across views. However, because the confidence score of fused boxes after WBF is an average of high-confidence and low-confidence inputs, their final confidence may be lower than that of the strongest single-view prediction. As a result, these fused boxes may not appear among the top-ranked outputs, but they become valuable when the confidence threshold is relaxed, thereby recovering more true positives and improving recall under occlusion. To illustrate this mechanism, Figure~\ref{fig:wbf-confidence} shows an example where a car is detected with 0.09 confidence in View~1 and 0.36 in View~2. After WBF, the fused detection reaches 0.15 confidence, which is higher than the weaker input but lower than the stronger one.

\paragraph{Three-view Scalability}\mbox{}\\
After establishing the effectiveness of WBF in the two-view setting, we further evaluate its scalability to three-view fusion. Table~\ref{tab:three-view-occlusion} shows that three-view decision-level fusion consistently outperforms two-view fusion in most occlusion triplets.

The benefits of three-view fusion are evident under both asymmetric and symmetric occlusion. In asymmetric occlusion, the severely occluded view is strongly compensated by the additional information from the less occluded views. For example, in the (30\%, 30\%, 50\%) setting, the third view improves by +0.3827 mAP over the baseline, and in the (30\%, 50\%, 50\%) setting, the second and third views gain +0.3413 mAP and +0.3444 mAP, respectively. In the symmetric (50\%, 50\%, 50\%) case, three-view fusion also outperforms both the two-view and single-view baselines, achieving improvements of +0.1538, +0.2124, and +0.1815 mAP across the three views. These results demonstrate that three-view collaborative inference remains effective and robust even under severe occlusion.

\begin{table}[t]
\centering
\caption{Comparison of baseline, two-view, and three-view fusion across occlusion settings.}
\label{tab:three-view-occlusion}
\resizebox{\columnwidth}{!}{%
\begin{tabular}{@{}llccccc@{}}
\toprule
Occ. triplet & View & Baseline & 2-view & $\Delta$ (2-view) & 3-view & $\Delta$ (3-view) \\
\midrule
30\%, 30\%, 30\% & View 1 & 0.6254 & 0.6739 & +0.0485 & 0.7226 & +0.0972 \\
30\%, 30\%, 30\% & View 2 & 0.5985 & 0.7192 & +0.1207 & 0.7685 & +0.1700 \\
30\%, 30\%, 30\% & View 3 & 0.6447 & 0.6447* & +0.0000* & 0.7501 & +0.1054 \\
\midrule
30\%, 30\%, 50\% & View 1 & 0.6254 & 0.6739 & +0.0485 & 0.7159 & +0.0905 \\
30\%, 30\%, 50\% & View 2 & 0.5985 & 0.7192 & +0.1207 & 0.7380 & +0.1395 \\
30\%, 30\%, 50\% & View 3 & 0.2683 & 0.2683* & +0.0000* & 0.6510 & +0.3827 \\
\midrule
30\%, 50\%, 50\% & View 1 & 0.6254 & 0.6661 & +0.0407 & 0.6839 & +0.0585 \\
30\%, 50\%, 50\% & View 2 & 0.2851 & 0.5770 & +0.2919 & 0.6264 & +0.3413 \\
30\%, 50\%, 50\% & View 3 & 0.2683 & 0.2683* & +0.0000* & 0.6127 & +0.3444 \\
\midrule
50\%, 50\%, 50\% & View 1 & 0.2965 & 0.3691 & +0.0726 & 0.4503 & +0.1538 \\
50\%, 50\%, 50\% & View 2 & 0.2851 & 0.3900 & +0.1049 & 0.4975 & +0.2124 \\
50\%, 50\%, 50\% & View 3 & 0.2683 & 0.2683* & +0.0000* & 0.4498 & +0.1815 \\
\bottomrule
\end{tabular}%
}
\\[2pt]
\footnotesize{All values report mAP@0.5. $\Delta$ indicates the performance change relative to the baseline. * For two-view fusion, the third view does not participate; its single-view detection score is therefore used for comparison.}
\end{table}

\begin{figure}[t]
\centering
\includegraphics[width=0.9\columnwidth]{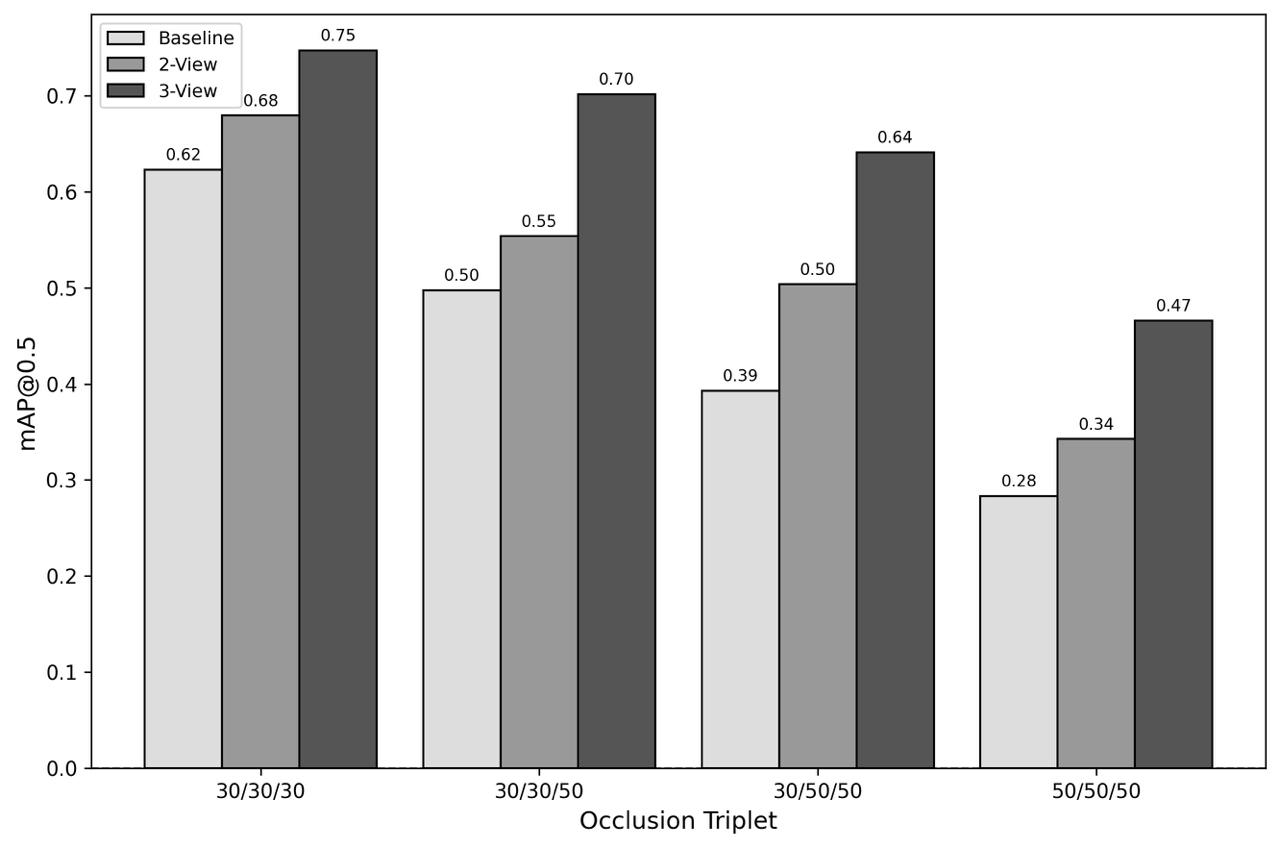}
\caption{Two-view versus three-view fusion accuracy across occlusion triplets.}
\Description{Comparison plot of two-view and three-view fusion accuracy across several occlusion triplets, showing that adding a third view usually improves detection performance.}
\label{fig:three-view-triplets}
\end{figure}

In Figure~\ref{fig:three-view-triplets}, we report the average mAP of single-view, two-view, and three-view fusion under four occlusion configurations. The bar chart highlights the consistent trend that three-view fusion outperforms both the single-view baseline and two-view fusion across all settings. The advantage is most pronounced in triplets that include a heavily occluded view (e.g., 50\%), where the additional view provides complementary information that helps recover detections under severe occlusion. This demonstrates the robustness of three-view fusion and its effectiveness in mitigating severe occlusion.

\paragraph{Wi-Fi Communication Overhead}\mbox{}\\
While three-view fusion improves accuracy, it inevitably incurs additional communication overhead. We therefore first evaluate Wi-Fi transmission costs in a controlled setup to examine the trade-off between the added communication burden and the performance gains achieved through fusion. At this stage, peer exchange is emulated between one MCU and a host endpoint representing a second node; Section~\ref{subsec:hardware-wifi} later activates the same Wi-Fi path on two physical boards. Detected bounding boxes are transmitted using UDP, since real-time delivery is prioritised over guaranteed reliability (packet loss only results in missed accuracy improvements, not incorrect predictions).

As shown in Table~\ref{tab:wifi-overhead}, packet size scales approximately linearly with the bounding box limit. When transmitting 80 bounding boxes, the payload exceeded the practical Ethernet Wi-Fi MTU limit in our implementation (1500 bytes)~\cite{ref36}, leading to fragmentation effects and consistent packet loss. Therefore, a limit of 60 bounding boxes represents the practical upper bound for reliable communication in the current setup. These findings suggest that decision-level fusion is more suitable than feature-level fusion in collaborative inference, as it maintains accuracy gains while keeping communication overhead within realistic network constraints.

\begin{table}[t]
\centering
\caption{On-device communication overhead via Wi-Fi.}
\label{tab:wifi-overhead}
\begin{tabular}{@{}ccc@{}}
\toprule
Bounding box limit per image & Packet size (bytes) & Lost packets \\
\midrule
10 & 262 & 0 \\
20 & 472 & 0 \\
40 & 892 & 0 \\
60 & 1312 & 0 \\
80 & 1732 & 100* \\
\bottomrule
\end{tabular}
\\[2pt]
\footnotesize{* Consistent packet loss observed after exceeding the practical Wi-Fi MTU limit in our implementation.}
\end{table}

\begin{figure}[t]
\centering
\includegraphics[width=0.9\columnwidth]{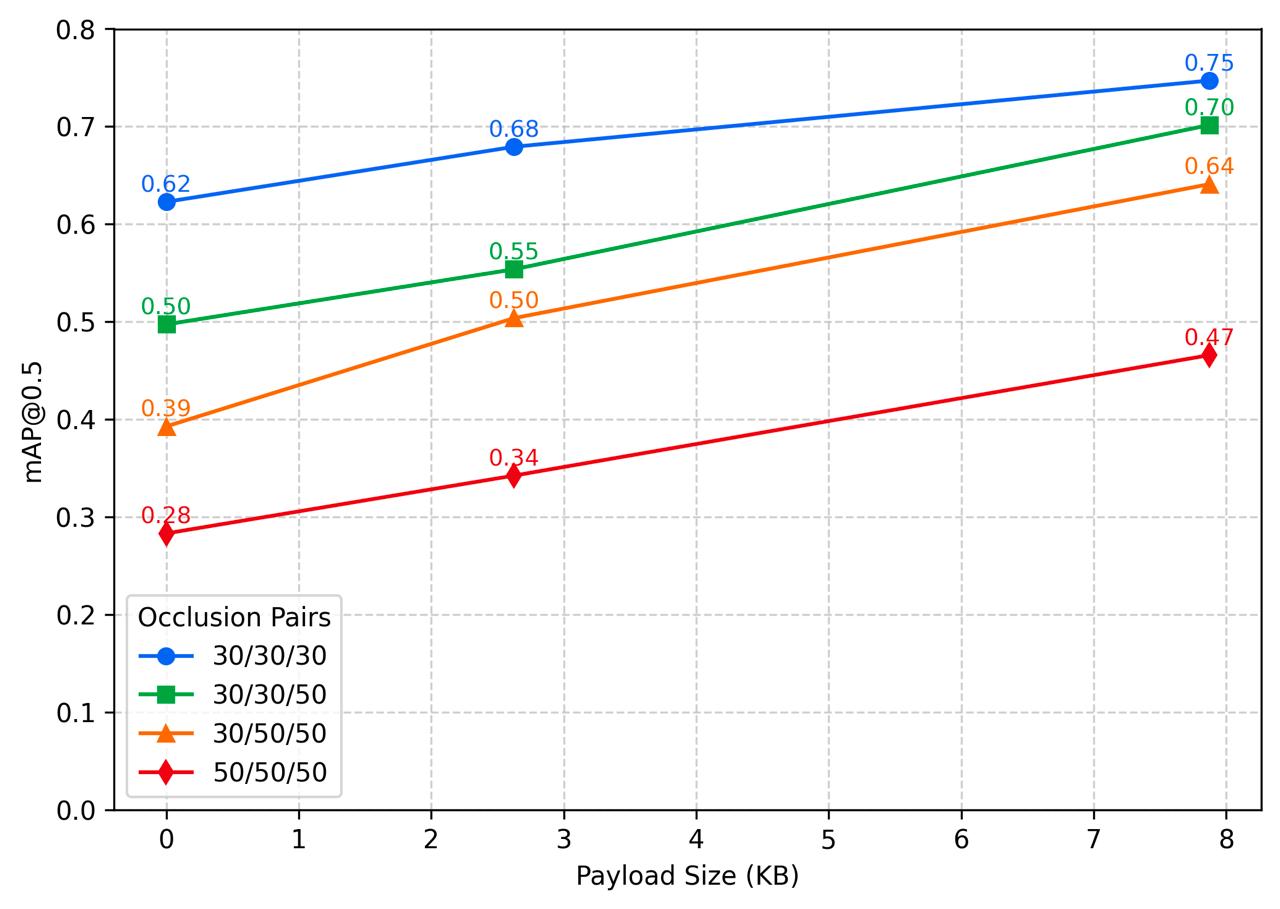}
\caption{Accuracy versus communication cost for two-view and three-view fusion.}
\Description{Plot relating detection accuracy to communication cost for two-view and three-view collaborative fusion, showing higher accuracy can be obtained at the cost of more transmitted data.}
\label{fig:accuracy-comm-cost}
\end{figure}

We further combine the measured payload per bounding-box exchange with the detection performance of two-view and three-view fusion (Figure~\ref{fig:accuracy-comm-cost}). The extra view in three-view fusion improves accuracy by approximately 0.07--0.15 mAP depending on the occlusion setting, but incurs about 6 KB of added communication overhead. This overhead is computed from six communication exchanges in the ring structure, with each payload measuring 1312 bytes ($\approx$ 1.3 KB). This highlights a clear trade-off: while accuracy improves, communication cost must be carefully balanced against device and network constraints in practical deployments.

\subsection{Exploratory Decentralised Training Study}\label{subsec:results-dfl}

\begin{figure}[t]
\centering
\includegraphics[width=0.9\columnwidth]{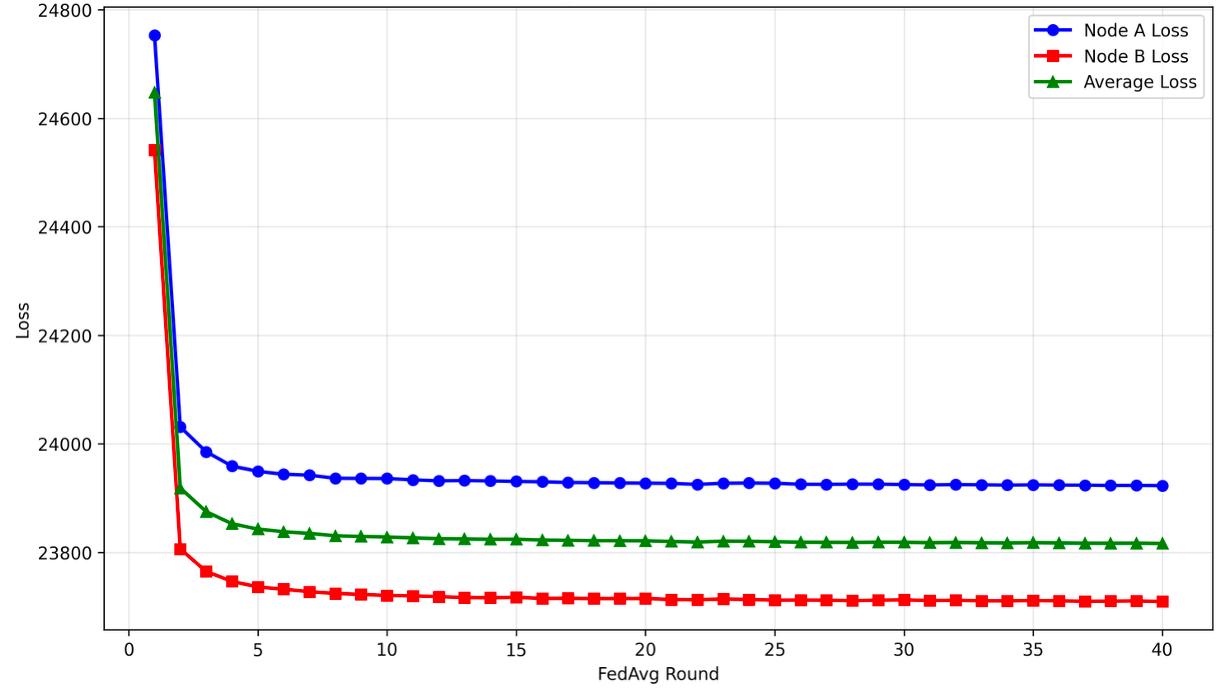}
\caption{Training loss curve of FedAvg over successive rounds.}
\Description{Line plot of FedAvg training loss across communication rounds for the two participating nodes and their average, showing a rapid initial drop followed by stabilisation.}
\label{fig:fedavg-loss}
\end{figure}

As a secondary exploratory experiment, we implemented a preliminary DFL framework to test whether decentralised training remains numerically stable in this setting. The results, shown in Figure~\ref{fig:fedavg-loss}, indicate that this setup reaches a stable training regime under FedAvg-style peer averaging. Training losses decrease rapidly in the first few rounds and then plateau.

However, the absolute loss values remain high ($\approx$ 23{,}800), suggesting that this stability does not yet translate into meaningful performance improvements. This outcome likely results from the non-iid data distribution, as different views were assigned to different devices. Accordingly, this section should be interpreted as a feasibility result on training stability rather than as evidence of deployment-ready adaptation gains.

\subsection{Limitations of the Experimental Evaluation}\label{subsec:limitations}

Before moving to real-hardware deployment, it is important to separate the controlled evaluation in this section from the end-to-end system validation that follows. Sections~\ref{subsec:results-pretraining} and~\ref{subsec:results-collab} establish the main algorithmic case for quantised detection and collaborative inference, whereas Section~\ref{subsec:results-dfl} serves only as an exploratory side study on decentralised training stability. Even so, several limitations shape the interpretation of the overall results.

\begin{itemize}
\item \textbf{Deployment gap in the controlled study:} The collaborative experiments above establish algorithmic behaviour and communication scaling, but they do not by themselves demonstrate simultaneous live operation on two fully independent camera-equipped boards. Section~\ref{sec:real-hardware} addresses this gap in two stages: first through a host-assisted USB-relay baseline, and then through a fully autonomous Wi-Fi peer-to-peer deployment.
\item \textbf{Pre-training gap:} The lightweight backbone relied on pre-trained MCUNet weights from prior studies, without large-scale pre-training tailored to the detection task. This may constrain accuracy and generalisability.
\item \textbf{Dataset scope:} Collaborative inference experiments were deliberately conducted on a small CO3D subset focused on the car category (approximately 100 images), because the study targets occlusion-robust object detection within a single class rather than broad multi-class recognition. This is sufficient for proof-of-concept evaluation, but the small scale and single-class setting limit the broader applicability of the findings.
\item \textbf{Simplified feature-level fusion:} Feature fusion was implemented using only batch normalisation alignment, which provided a baseline but omitted more advanced mechanisms, such as feature calibration. This may explain the instability observed in feature-level fusion.
\item \textbf{Exploratory decentralised learning baseline:} In the DFL stage, we implement FedAvg as a baseline, without exploring more advanced approaches for handling non-iid data. This narrows the conclusions regarding decentralised learning in heterogeneous edge environments.
\end{itemize}

Taken together, these limitations motivate two next steps. The first is immediate and practical: validating the complete pipeline on real hardware across both host-assisted and host-free deployments, which is presented in Section~\ref{sec:real-hardware}. The second is longer-term: improving feature alignment, extending evaluation to larger multi-class datasets, and exploring stronger decentralised learning algorithms.

\section{Real Hardware Deployment}\label{sec:real-hardware}

\subsection{Motivation}\label{subsec:hardware-motivation}
The earlier sections establish algorithmic feasibility for the MCUNet--YOLOv2 pipeline: Section~\ref{subsec:results-pretraining} quantifies the effects of quantisation and on-device inference, whereas Section~\ref{subsec:results-collab} shows that WBF improves detection under asymmetric occlusion. However, those results still leave one systems question open: can the full detection-and-fusion pipeline run end-to-end across two physically independent boards, each with its own live camera feed?

This section answers that question in two stages. We first establish a host-assisted USB-relay baseline that validates simultaneous live sensing, on-device detection, and cross-board fusion on two physical Coral Dev Board Micros. We then activate the firmware's Wi-Fi path on both boards, remove the host PC from the loop, and execute WBF directly on each board's M7 core. The contribution here is therefore not new fusion logic (the WBF procedure is unchanged from Section~\ref{subsec:results-collab}), but a systems progression from controlled two-board validation to fully autonomous peer-to-peer deployment.

\subsection{USB-Relay Hardware Setup}\label{subsec:hardware-setup}
For the initial live deployment, two Coral Dev Board Micros are positioned at different viewpoints of the same target scene, each connected to the host PC by a separate USB cable. The boards operate fully independently: there is no direct communication between them, and neither board is aware of the other's detections. All fusion logic runs on the host.

Each board uses its integrated Himax color camera, configured here to output $160 \times 160$ RGB frames. The two cameras are oriented to provide complementary views of the target, deliberately replicating the asymmetric occlusion scenario studied in Section~\ref{subsec:results-collab}: one board captures a broadside view, while the other captures a partial foreshortened profile. This mirrors the $(30\%, 50\%)$ occlusion pair in Table~\ref{tab:two-view-occlusion}, the setting where WBF delivered its largest gain (+0.2736 mAP).

\subsection{USB-Relay Firmware Architecture}\label{subsec:hardware-firmware}
The firmware runs a single-threaded inference loop on the M7 core. Each iteration proceeds through five stages:

\begin{enumerate}
\item \textbf{Frame capture.} The integrated Himax camera capture pipeline fills a $160 \times 160 \times 3$ NHWC buffer.
\item \textbf{Inference.} The TFLM interpreter runs \texttt{Invoke()} on the INT8 model loaded from the SD card.
\item \textbf{Post-processing.} The $5 \times 5 \times 30$ output tensor is decoded using a sigmoid on objectness and class score, an exponential on box dimensions, anchor-box coordinate recovery, and score-threshold filtering at 0.40.
\item \textbf{Detection serialisation.} Surviving detections are packed as binary \texttt{WireDet} structs and sent over USB serial to the host relay.
\item \textbf{Camera snapshot (optional).} When the compile-time flag \texttt{SEND\_CAMERA\_FRAMES} is enabled, the board additionally transmits the current frame subsampled to $80 \times 80$, gated on the condition that at least one detection survives.
\end{enumerate}

Section~\ref{subsec:hardware-wifi} describes the firmware modifications that replace this serial exchange path with direct Wi-Fi peer communication.

Both packet types share a two-byte magic preamble (\texttt{0xFE} followed by a type byte). The wire protocol is summarised in Table~\ref{tab:wire-protocol}.

\begin{table}[t]
\centering
\caption{Binary packet types used in the USB serial relay.}
\label{tab:wire-protocol}
\small
\begin{tabular}{@{}p{0.19\columnwidth}p{0.14\columnwidth}p{0.57\columnwidth}@{}}
\toprule
Packet type & Second byte & Payload \\
\midrule
Detection & \texttt{0xED} & \texttt{board\_id}, \texttt{num\_dets}, and $N \times$ 10-byte \texttt{WireDet} records \\
Camera snapshot & \texttt{0xEE} & \texttt{board\_id}, \texttt{img\_w}, \texttt{img\_h}, \texttt{num\_dets}, $N \times$ 10-byte \texttt{WireDet} records, followed by \texttt{img\_w} $\times$ \texttt{img\_h} $\times 3$ raw RGB bytes \\
\bottomrule
\end{tabular}
\end{table}

Each \texttt{WireDet} encodes five \texttt{uint16} fields: \texttt{x1}, \texttt{y1}, \texttt{x2}, \texttt{y2} at pixel $\times 4$ scale, and \texttt{score} at score $\times 10{,}000$. This gives sub-pixel coordinate resolution and four decimal places of score precision in 10 bytes per detection.

The camera snapshot is 2:1 subsampled rather than sent as the full $160 \times 160$ frame. Every other pixel in both dimensions is taken, producing an $80 \times 80$ image (19{,}200 bytes) that preserves the spatial structure of the detections while reducing the raw pixel payload to one quarter of the full frame, i.e., a 75\% reduction in transfer size. The snapshot is transmitted only when detections are present, avoiding unnecessary serial traffic on empty frames.

\subsection{Host-Side Relay}\label{subsec:hardware-relay}
A high-level algorithmic description of the host relay is given in Appendix~\ref{app:host-relay}. The relay runs two threads on the host, one per USB serial port, and performs three functions.

\paragraph{Packet parsing.}\mbox{}\\
Each thread scans the byte stream for the \texttt{0xFE} magic byte, then branches on the second byte to dispatch to the detection or snapshot handler. Non-magic bytes are printed as debug text, allowing firmware \texttt{Serial.print()} messages to pass through transparently.

\paragraph{WBF fusion.}\mbox{}\\
After either board delivers a detection packet, the relay runs WBF across both boards' most recent detection lists. The algorithm and parameters are identical to those used in Section~\ref{subsec:results-collab}: IoU threshold 0.45, score threshold 0.40, and fused score computed as the arithmetic mean of the cluster. Fusion results are printed to the terminal alongside each board's individual detection count, enabling direct comparison of single-view and fused outputs in real time.

\paragraph{Camera frame reconstruction.}\mbox{}\\
The snapshot handler accumulates the raw RGB payload using a loop-based reader (\texttt{read\_exactly}; Algorithm~\ref{alg:relay-read-exactly}) that continues until the full 19{,}200 bytes have been received. This was a critical engineering requirement: at 115{,}200 baud, \texttt{pyserial}'s default per-call read timeout of 0.5 seconds expires after approximately 5{,}760 bytes, less than a third of the full payload, causing systematic truncation. The loop-based reader resolves this by accumulating chunks across as many reads as necessary, with a 30-second deadline. Once complete, the payload is decoded with Pillow, annotated with bounding boxes and score labels, and saved to \texttt{captures/board<id>\_<frame>.png}.

\subsection{USB-Relay Performance}\label{subsec:hardware-performance}
\paragraph{Inference latency.}\mbox{}\\
The per-frame invoke time on the M7 core is consistently $2{,}403{,}000$--$2{,}407{,}000$ $\mu$s (approximately 2.4 seconds), corresponding to roughly 0.41 FPS. This is notably lower than the 3{,}197 ms reported in Table~\ref{tab:on-device-latency}. The difference is attributable to the quantisation pipeline: the earlier evaluation uses the standard TFLite post-training quantisation converter, whereas this deployment uses \texttt{onnx2tf}'s per-channel INT8 export path, which generates more efficient operator sequences for depthwise-separable convolutions on the M7 core. The result confirms that quantisation tool choice has a material impact on on-device latency beyond the accuracy effects reported in Table~\ref{tab:quantised-vs-original}.

\paragraph{Memory usage.}\mbox{}\\
The TFLM arena required at runtime is 411 KB, compared with 759.2 KB reported in Table~\ref{tab:on-device-memory}. Again, this reflects the difference in quantisation pipeline: \texttt{onnx2tf}'s per-channel quantisation produces smaller intermediate activation tensors, reducing peak arena consumption by approximately 46\%. Both deployments comfortably satisfy the 1 MB SRAM constraint.

\paragraph{Host processing.}\mbox{}\\
Detection packet decoding on the host takes 55--59 $\mu$s. WBF fusion across the two boards' detections takes 0--34 $\mu$s. Total host processing overhead per inference cycle is therefore under 0.1 ms, which is negligible relative to the 2.4-second on-device cycle.

\paragraph{Serial bandwidth.}\mbox{}\\
At 115{,}200 baud, a detection-only packet for a single detection is 14 bytes (2-byte preamble, 2-byte header, and one 10-byte \texttt{WireDet}) and transfers in under 2 ms. A camera snapshot packet for an $80 \times 80$ frame is approximately 19{,}220 bytes, requiring about 1.67 seconds (roughly 70\% of a full inference cycle). The snapshot path is therefore reserved for diagnostic use (\texttt{SEND\_CAMERA\_FRAMES=0} in production builds).

A consolidated comparison of the two real-hardware deployments is deferred to Table~\ref{tab:hardware-summary} at the end of this section.

\subsection{USB-Relay Detection Results}\label{subsec:hardware-results}
\paragraph{Score range.}\mbox{}\\
Across the live test session, Board~1 (broadside view) produced detection scores of 0.40--0.685, whereas Board~2 (foreshortened view) produced scores of 0.40--0.53. The asymmetry is consistent with the $(30\%, 50\%)$ occlusion pair in Table~\ref{tab:two-view-occlusion}: the better-aligned view produces higher-confidence predictions. The absolute score values are lower than the CO3D evaluation figures from Section~\ref{subsec:results-collab}, which is expected because the CO3D frames are high-resolution photographs, whereas the integrated Himax camera at $160 \times 160$ produces a noisier, lower-contrast signal.

\paragraph{Complementary coverage.}\mbox{}\\
Multiple frames in the live session show one board producing zero detections (below threshold) while the other fires confidently. In these frames, the fused result correctly inherits the surviving detection. This behaviour is the real-hardware analogue of the asymmetric occlusion advantage observed in Section~\ref{subsec:results-collab}: the multi-view configuration recovers detections that a single board would otherwise miss.

\paragraph{Bounding box quality.}\mbox{}\\
In representative annotated captures from each board, Board~1's boxes are tightly localised around the vehicle body, whereas Board~2's boxes are wider and slightly shifted, reflecting the steeper viewing angle. Both are correctly localised, and no false positives were observed in the captured set. When both boards fire simultaneously, the WBF output produces a consensus box geometrically between the two views, matching the confidence-averaging mechanism illustrated in Figure~\ref{fig:wbf-confidence}.

\subsection{USB-Relay Discussion}\label{subsec:hardware-discussion}
The USB-relay deployment reproduces the qualitative findings of Section~\ref{subsec:results-collab} on real hardware. WBF fusion over two independent camera views improves coverage in the asymmetric case (the primary finding of Section~\ref{subsec:results-collab}), and the model runs reliably within the memory budget established in Section~\ref{subsec:results-pretraining}. The key quantitative differences (lower latency and smaller arena size) are attributable to the \texttt{onnx2tf} quantisation pipeline rather than any change to the model architecture.

Two aspects of this initial live deployment go beyond the earlier algorithmic evaluation. First, the camera snapshot feature provides direct visual evidence that the model is detecting the correct region of the physical scene rather than responding to dataset-specific cues (a validation that cannot be obtained from mAP values alone). Second, the binary serial protocol establishes an engineering baseline for multi-board deployments: detection packets are compact (14 bytes for a single detection), impose negligible latency overhead, and are robust to the byte-level framing issues that can arise from mixing binary and text output on a shared serial channel.

The USB-relay deployment establishes a dependable engineering baseline, but it still leaves one architectural limitation: the host PC remains in the loop for inter-board exchange and WBF execution. Subsection~\ref{subsec:hardware-wifi} removes this dependency by activating the firmware's Wi-Fi peer-to-peer path on both boards. Even after that transition, inference throughput remains the dominant system bottleneck. At 0.41 FPS in the USB-relay baseline, the system is suitable for stationary or slow-moving targets, but not for vehicles in motion at realistic speeds. Subsection~\ref{subsec:hardware-future} outlines the remaining path to higher throughput.

\subsection{Wi-Fi Peer-to-Peer Deployment}\label{subsec:hardware-wifi}
\paragraph{Overview.}\mbox{}\\
The USB-relay baseline above validates multi-board live inference, but it still depends on a tethered host. This subsection reports the next systems step: activation of the \texttt{CORAL\_MICRO\_WIFI} path on both Coral Dev Board Micro units, connection to a shared IEEE 802.11n access point, and live collaborative inference with no host PC in the loop. Each board captures frames from its integrated Himax camera, performs on-device detection, exchanges detection packets over Wi-Fi, and executes WBF locally on its M7 core. The host PC is absent during operation, and no relay script runs. A high-level algorithmic description of this on-device Wi-Fi path is given in Appendix~\ref{app:wifi-peer}.

\paragraph{Firmware modifications.}\mbox{}\\
The USB-relay path requires two main changes to operate wirelessly. First, detection packets are redirected from USB serial to a UDP socket. Each board opens a UDP socket on port 5005, enables the \texttt{SO\_BROADCAST} socket option, and transmits to \texttt{255.255.255.255}, eliminating the need for static IP configuration: any board joining the local subnet becomes a potential fusion peer. Second, the receiver discards self-originated broadcasts by checking the two-byte packet header (\texttt{board\_id}, \texttt{num\_dets}) before decoding. WBF, previously executed by the host relay, is re-implemented in C++ on the M7 core and invoked immediately after \texttt{RecvDets()} returns. The fusion parameters remain identical to the rest of this study: IoU threshold 0.45, score threshold 0.40, and fused score computed as the arithmetic mean of the contributing per-view scores, as in Algorithm~\ref{alg:wbf}. The \texttt{WireDet} encoding is unchanged: five \texttt{uint16} fields encode \texttt{x1}, \texttt{y1}, \texttt{x2}, \texttt{y2} at pixel $\times 4$ scale and \texttt{score} at score $\times 10{,}000$, giving 10 bytes per detection. A typical two-detection UDP message therefore consists of a 2-byte header plus 20 bytes of detection data, i.e., 22 bytes at the firmware packet level. By contrast, the equivalent USB stream packet with two detections is 24 bytes because it includes the 2-byte \texttt{0xFE 0xED} preamble required for byte-stream framing.

Table~\ref{tab:usb-vs-wifi-protocol} contrasts the protocol stack of the USB-relay and Wi-Fi peer-to-peer deployments.

\begin{table}[t]
\centering
\caption{Protocol comparison between the USB-relay baseline and the Wi-Fi peer-to-peer deployment.}
\label{tab:usb-vs-wifi-protocol}
\small
\begin{tabular}{@{}p{0.24\columnwidth}p{0.30\columnwidth}p{0.30\columnwidth}@{}}
\toprule
Property & USB relay & Wi-Fi peer-to-peer \\
\midrule
Transport & USB serial, 115{,}200 baud & UDP/IPv4 broadcast over the CYW43455 \\
Destination & Board $\rightarrow$ host PC (wired, unicast) & \texttt{255.255.255.255}:5005 \\
Packet framing & \texttt{0xFE} magic byte + type byte & Datagram boundary preserved by UDP \\
Header & \texttt{board\_id} + \texttt{num\_dets} (2 bytes) & \texttt{board\_id} + \texttt{num\_dets} (2 bytes) \\
Per-detection payload & 10-byte \texttt{WireDet} & 10-byte \texttt{WireDet} \\
Packet bytes, 2 detections & 24 bytes & 22 bytes \\
Self-filtering & Not required (separate host ports) & \texttt{board\_id} check in firmware \\
WBF execution & Host PC, Python, 0--34 $\mu$s & On-device, C++, M7 core, $<1$ $\mu$s \\
IP address management & Not required & Not required (broadcast) \\
Host PC required & Yes & No \\
\bottomrule
\end{tabular}
\\[2pt]
\footnotesize{USB packet counts include the 2-byte \texttt{0xFE 0xED} framing preamble required for byte-stream transport; the UDP packet does not require this extra framing.}
\end{table}

\paragraph{Performance.}\mbox{}\\
Table~\ref{tab:usb-vs-wifi-performance} compares the measured runtime of the two deployments.

\begin{table}[t]
\centering
\caption{Runtime comparison between the USB-relay baseline and the Wi-Fi peer-to-peer deployment.}
\label{tab:usb-vs-wifi-performance}
\small
\begin{tabular}{@{}p{0.30\columnwidth}p{0.17\columnwidth}p{0.18\columnwidth}p{0.15\columnwidth}@{}}
\toprule
Metric & USB relay & Wi-Fi peer-to-peer & $\Delta$ \\
\midrule
Per-cycle latency & $\sim$2.403 s & $\sim$2.774 s & +371 ms (+15.4\%) \\
Effective frame rate & $\sim$0.41 FPS & $\sim$0.36 FPS & --0.05 FPS \\
Packet bytes (2 detections) & 24 & 22 & --2 bytes \\
Packet transfer time & $<2.1$ ms (serial) & $\sim$70 $\mu$s (UDP TX) & $\approx$--2.0 ms \\
Peer receive time & N/A (host-mediated) & 4--8 $\mu$s (UDP RX) & N/A \\
WBF latency & 0--34 $\mu$s (host Python) & $<1$ $\mu$s (on-device C++) & N/A \\
TFLM arena & 411 KB & 411 KB & 0 \\
Host PC required & Yes & No & N/A \\
\bottomrule
\end{tabular}
\end{table}

The additional 371 ms of per-cycle latency arises from two sources. The CYW43455 wireless driver executes interrupt service routines on the same M7 bus that services TFLM operator dispatch, introducing variable contention during inference. In addition, the firmware blocks on \texttt{RecvDets()} until the peer's UDP datagram arrives, introducing a synchronisation wait bounded by the phase offset between the two boards' independent inference clocks. Both effects are structural consequences of running the Wi-Fi stack concurrently with TFLM on a single core, rather than properties of UDP itself: the actual over-the-air transmission completes in approximately 70 $\mu$s, more than an order of magnitude faster than the equivalent 115{,}200-baud serial transfer. Despite the overhead, the system operates at approximately 0.36 FPS, which remains adequate for the stationary and slow-moving targets studied in this deployment.

\paragraph{Energy analysis.}\mbox{}\\
Table~\ref{tab:wifi-energy} reports firmware-side energy estimates for Wi-Fi mode. The firmware computes these values as $E = P \times t$ using the measured durations of inference, UDP transmit, and UDP receive together with nominal power constants embedded in the firmware. These are simplified estimates rather than external power measurements, so the absolute $\mu$J values should be interpreted as order-of-magnitude indicators rather than instrumented board-level measurements.

\begin{table}[t]
\centering
\caption{Estimated per-cycle energy breakdown in the Wi-Fi peer-to-peer deployment.}
\label{tab:wifi-energy}
\small
\begin{tabular}{@{}p{0.34\columnwidth}p{0.18\columnwidth}p{0.22\columnwidth}@{}}
\toprule
Component & Estimated energy ($\mu$J) & Share of total \\
\midrule
TFLM inference (M7 core) & $\sim$222{,}000 & $\sim$99.997\% \\
UDP transmit (22-byte packet) & $\sim$7 & $\sim$0.003\% \\
UDP receive (22-byte packet) & $\sim$0.25 & $<0.001\%$ \\
Total per cycle & $\sim$222{,}007 & 100\% \\
\bottomrule
\end{tabular}
\\[2pt]
\footnotesize{Estimated in firmware using nominal power constants of 80 mW for CPU-active inference, 100 mW for Wi-Fi transmit, and 50 mW for Wi-Fi receive.}
\end{table}

The collaborative communication layer is therefore negligible in relative terms. Under the nominal firmware constants above, transmitting a complete detection packet costs approximately 7 $\mu$J, roughly 1/31{,}700 of the energy consumed by a single TFLM inference pass, and the receive path costs approximately 0.25 $\mu$J. Even if the wireless power constants are revised upward, communication remains a very small fraction of the cycle energy because transmit and receive occupy only tens of microseconds per cycle, whereas inference lasts approximately 2.774 seconds.

\paragraph{Detection results.}\mbox{}\\
The complementary coverage behaviour characterised in Section~\ref{subsec:hardware-results} is reproduced faithfully in Wi-Fi mode. Table~\ref{tab:wifi-session} summarises one representative autonomous session from Board~2's perspective. The session spans 301.9 s and 108 frames, with an average frame interval of 2.821 s (0.354 FPS). Within that session, 14 frames are recovery cases in which Board~2 produces no surviving detection while Board~1 supplies a peer detection, and fused output is present in 61 frames overall. Relative to Board~2 alone, collaborative fusion increases frame-level coverage from 47 to 61 frames, a gain of 14 frames (+29.8\%).

\begin{table}[t]
\centering
\caption{Representative autonomous Wi-Fi session summary, reported from Board~2's perspective.}
\label{tab:wifi-session}
\small
\begin{tabular}{@{}p{0.55\columnwidth}p{0.25\columnwidth}@{}}
\toprule
Metric & Value \\
\midrule
Total frames & 108 (F1--F108) \\
Session duration & 301.9 s (5 min 2 s) \\
Average frame interval & 2.821 s (0.354 FPS) \\
Both boards silent (own = 0, peer = 0) & 47 (43.5\%) \\
Board~2 only (own $>$ 0, peer = 0) & 28 (25.9\%) \\
Board~1 only -- recovery (own = 0, peer $>$ 0) & 14 (13.0\%) \\
Both active (own $>$ 0, peer $>$ 0) & 19 (17.6\%) \\
Any fused output (fused $>$ 0) & 61 (56.5\%) \\
Single-view coverage (Board~2 alone) & 47 frames \\
Collaborative coverage (fused) & 61 frames \\
Gain from fusion & +14 frames (+29.8\%) \\
\bottomrule
\end{tabular}
\end{table}

Score distributions are also consistent with the USB-relay baseline: Board~1 produces confidence scores in the range 0.40--0.68, and Board~2 in the range 0.40--0.53. The asymmetry reflects the same $(30\%, 50\%)$ occlusion geometry described earlier in this section, with the broader-angle view yielding higher-confidence predictions. The change in communication medium introduces no observable accuracy regression: fused scores remain the arithmetic mean of the per-view inputs, identical to the host-relay implementation. No false positives were observed during the recorded live session.

\paragraph{Discussion.}\mbox{}\\
The Wi-Fi deployment addresses three practical limitations of the USB-relay baseline. The first is physical autonomy. The system operates with no host PC, no relay script, and no USB tether. Two boards, each powered independently, capture frames, run inference, exchange detections over UDP, and fuse results on-device within a single inference cycle. This closes the deployment gap identified in Section~\ref{subsec:limitations}.

The second is zero-configuration networking. UDP broadcast to \texttt{255.255.255.255} removes the need for manual IP management, so DHCP-assigned addresses, board reboots, and routine network changes remain transparent to the protocol. It also simplifies future extension to larger peer groups, although additional application-layer logic would be required to coordinate fusion rounds beyond the current two-board setting.

The third is that WBF runs on the sensor node itself. Moving fusion from the host (Python, tens of microseconds of host CPU time) to the M7 core (C++, sub-microsecond) is not primarily about latency, because both are negligible relative to the multi-second inference cycle. It is mainly an architectural shift. Fusion logic now resides on the hardware that captures the scene, eliminating the host as a point of failure and enabling deployment where no persistent compute infrastructure is available.

The main cost of this architecture is the 15.4\% increase in per-cycle latency relative to the USB-relay baseline, attributable to Wi-Fi stack contention and cross-board synchronisation on the shared M7 core. This overhead would become proportionally smaller if inference were offloaded to the Edge TPU co-processor: with throughput above 5 FPS, the M7 core would handle only pre- and post-processing, and the CYW43455 driver's interrupt load would occupy a much smaller fraction of the total cycle. For this system, the Wi-Fi peer-to-peer design is the more deployable endpoint, and Edge TPU acceleration is the next engineering step toward real-time collaborative inference at the edge.

\subsection{Remaining Future Work}\label{subsec:hardware-future}
After the autonomous Wi-Fi deployment above, two engineering directions remain central to real-time collaborative inference on this platform.

\paragraph{Edge TPU acceleration.}\mbox{}\\
The Coral Dev Board Micro includes a Google Edge TPU co-processor that was not used in either deployment reported here. Enabling it for the backbone's depthwise-separable convolutions (which dominate the inference FLOPs) is expected to reduce latency by roughly an order of magnitude, bringing throughput above 5 FPS and making the system viable for moving targets.

\paragraph{Timestamp synchronisation.}\mbox{}\\
In the Wi-Fi deployment, each board fuses the most recent peer detections without explicit timestamp alignment. At 0.36 FPS, the temporal offset between boards remains on the order of one inference cycle (approximately 2.8 seconds), which is acceptable for stationary targets. With Edge TPU acceleration bringing frame rates above 5 FPS, explicit timestamps in the detection-packet header would become necessary to avoid fusing temporally misaligned observations.

For convenience, Table~\ref{tab:hardware-summary} consolidates the key parameters of the two real-hardware deployments discussed in this section.

\begin{table}[H]
\centering
\caption{End-of-section summary of the two real-hardware deployments.}
\label{tab:hardware-summary}
\small
\begin{tabular}{@{}p{0.33\columnwidth}p{0.22\columnwidth}p{0.27\columnwidth}@{}}
\toprule
Parameter & USB relay & Wi-Fi peer-to-peer \\
\midrule
System role & Host-assisted baseline & Fully autonomous deployment \\
TFLM invoke latency & $\sim$2.403 s & $\sim$2.774 s \\
Effective frame rate & $\sim$0.41 FPS & $\sim$0.36 FPS \\
TFLM arena & 411 KB & 411 KB \\
Quantisation pipeline & \texttt{onnx2tf} per-channel INT8 & \texttt{onnx2tf} per-channel INT8 \\
Communication medium & USB serial (115{,}200 baud) & UDP/IPv4 broadcast over CYW43455 \\
Packet bytes, 2 detections & 24 & 22 \\
WBF execution & Host PC, Python & On-device, C++ on M7 \\
Camera frames captured & Yes (live Himax camera) & Yes (live Himax camera) \\
Host PC required & Yes & No \\
\bottomrule
\end{tabular}
\\[2pt]
\footnotesize{USB packet counts include the 2-byte \texttt{0xFE 0xED} framing preamble; Wi-Fi packet counts do not require it because UDP preserves packet boundaries.}
\end{table}

\section{Conclusion}

This work examined object detection on ultra--low-end MCUs under three constraints that often appear together in edge deployments: limited compute and memory, communication cost between cooperating devices, and reduced accuracy when objects are occluded or the local data distribution changes. The evaluated pipeline combines an MCUNet backbone, a YOLOv2 detection head, and TFLite quantisation, and is tested from controlled experiments through to real-hardware deployment.

In the algorithmic evaluation, MCUNet with YOLOv2 provides the best accuracy--efficiency balance among the lightweight baselines tested. Quantisation reduces model size and peak RAM usage by approximately 71\% and 83\%, respectively, with a small accuracy drop (--0.003 mAP). For collaborative inference, decision-level fusion via WBF improves detection under occlusion, with gains of up to +0.2736 mAP in the asymmetric two-view setting and up to +0.3827 mAP when extended to three views. Feature-level fusion is less stable in these experiments, and the three-view case shows that additional accuracy must be weighed against communication cost. The exploratory FedAvg-style DFL experiment remains numerically stable without central coordination, but its high absolute loss ($\approx$ 23{,}800) indicates that stronger adaptation methods and a more complete evaluation are needed before treating DFL as deployment-ready.

The hardware experiments move from a host-assisted USB-relay baseline to a Wi-Fi peer-to-peer deployment on two Coral Dev Board Micros using live Himax camera inputs. The USB relay provides a controlled baseline for packetisation, diagnostic capture, and host-side fusion. The Wi-Fi version removes the host and runs WBF directly on each board, with communication energy remaining small relative to inference. In a representative 301.9 s autonomous session with 108 frames, fused output is present on 61 frames compared with 47 for Board~2 alone, giving a frame-level coverage gain of +29.8\%. These two deployments reproduce the asymmetric-view benefit observed in the CO3D experiments and show that decision-level fusion can run on the target hardware without a host PC.

Overall, decision-level fusion is the most reliable option tested in this work for occlusion-aware edge perception. It does not require backbone retraining, avoids explicit view calibration, and remains compatible with tight memory and communication constraints. The main open issues are higher throughput, timestamp alignment at faster frame rates, evaluation on richer datasets, and stronger decentralised adaptation methods.

%% file: appendices.tex
\appendix
\numberwithin{equation}{section}
\numberwithin{algorithm}{section}
\clearpage
\section{Formulation}\label{app:formulation}
\subsection{YOLOv2 Loss Function}\label{app:yolov2-loss}
\begin{equation}
\mathcal{L} = \mathcal{L}_{\mathrm{coord}} + \mathcal{L}_{\mathrm{obj}} + \mathcal{L}_{\mathrm{cls}}
\label{eq:yolo-total-loss}
\end{equation}

The three components are given by:
\begin{equation}
\begin{aligned}
\mathcal{L}_{\mathrm{coord}} &= \lambda_{\mathrm{coord}} \sum_{i=0}^{S^2} \sum_{j=0}^{A} \mathbf{1}^{\mathrm{obj}}_{ij}\left[(x_{ij} - \hat{x}_{ij})^2 + (y_{ij} - \hat{y}_{ij})^2\right] \\
&\quad + \lambda_{\mathrm{coord}} \sum_{i=0}^{S^2} \sum_{j=0}^{A} \mathbf{1}^{\mathrm{obj}}_{ij}\left[(\sqrt{w_{ij}} - \sqrt{\hat{w}_{ij}})^2 + (\sqrt{h_{ij}} - \sqrt{\hat{h}_{ij}})^2\right]
\end{aligned}
\label{eq:yolo-coord-loss}
\end{equation}
\begin{equation}
\begin{aligned}
\mathcal{L}_{\mathrm{obj}} &= \lambda_{\mathrm{obj}} \sum_{i=0}^{S^2} \sum_{j=0}^{A} \mathbf{1}^{\mathrm{obj}}_{ij}\left(\mathrm{IOU}^{\mathrm{truth}}_{ij} - \hat{C}_{ij}\right)^2 \\
&\quad + \lambda_{\mathrm{noobj}} \sum_{i=0}^{S^2} \sum_{j=0}^{A} \mathbf{1}^{\mathrm{noobj}}_{ij}\left(0 - \hat{C}_{ij}\right)^2
\end{aligned}
\label{eq:yolo-obj-loss}
\end{equation}
\begin{equation}
\mathcal{L}_{\mathrm{cls}} = \sum_{i=0}^{S^2} \mathbf{1}^{\mathrm{obj}}_i \left(-\sum_{c \in \mathrm{classes}} \hat{p}_i(c) \log p_i(c)\right)
\label{eq:yolo-cls-loss}
\end{equation}

\paragraph{Notation.}
\begin{itemize}
\item $\mathbf{1}^{\mathrm{obj}}_{ij}$: indicator, equals 1 if anchor $j$ in cell $i$ is responsible for an object, else 0.
\item $\mathbf{1}^{\mathrm{noobj}}_{ij}$: indicator, equals 1 if anchor $j$ in cell $i$ is not responsible for any object, else 0.
\item $(x_{ij}, y_{ij}, w_{ij}, h_{ij})$: ground truth bounding box centre coordinates and dimensions (relative to the grid cell).
\item $(\hat{x}_{ij}, \hat{y}_{ij}, \hat{w}_{ij}, \hat{h}_{ij})$: predicted bounding box parameters.
\item $\mathrm{IOU}^{\mathrm{truth}}_{ij}$: intersection over union between predicted box $j$ in cell $i$ and the ground truth box.
\item $\hat{C}_{ij}$: predicted objectness score for anchor $j$ in cell $i$.
\item $\hat{p}_i(c)$ and $p_i(c)$: target indicator and predicted class probability for class $c$ in cell $i$, respectively.
\end{itemize}

\clearpage
\section{Hyperparameters}\label{app:hyperparameter}
\subsection{Hyperparameters of MCUNet with YOLOv2}\label{app:mcunet-hparams}
The training configuration is summarised as follows:
\begin{itemize}
\item Epochs: 160.
\item Optimiser: AdamW.
\item Base learning rate: $1 \times 10^{-3}$.
\item Weight decay: $1 \times 10^{-4}$.
\item Warm-up: 5 epochs.
\item Cosine annealing schedule with $\eta_{\min} = 0.05 \times$ base learning rate.
\item Loss weights: $\lambda_{\mathrm{coord}} = 5.0$, $\lambda_{\mathrm{obj}} = 5.0$, and $\lambda_{\mathrm{noobj}} = 0.5$.
\end{itemize}

\clearpage
\section{Algorithm}\label{app:algorithm}
\subsection{Weighted Boxes Fusion Algorithm}\label{app:wbf}
\begin{algorithm}[H]
\caption{Weighted Boxes Fusion (WBF) per Image}
\label{alg:wbf}
\begin{algorithmic}[1]
\Require Bounding boxes from View~1 and View~2, each represented as $(x_1, y_1, x_2, y_2, \mathrm{score}, \mathrm{class})$
\Ensure Final set of fused bounding boxes
\State Initialise an empty list $\mathrm{clusters}$ (each cluster contains a group of overlapping boxes)
\State Combine and sort all input boxes by confidence score in descending order
\For{each box $b$ in the sorted list}
    \State $\mathrm{matched} \gets \mathrm{False}$
    \For{each box $b'$ in each existing cluster $\mathrm{cluster}$}
        \If{$\mathrm{IoU}(b, b') > \mathrm{threshold}$}
            \State Add $b$ to the corresponding cluster
            \State $\mathrm{matched} \gets \mathrm{True}$
            \State \textbf{break}
        \EndIf
    \EndFor
    \If{not $\mathrm{matched}$}
        \State Create a new cluster containing $b$ and append it to $\mathrm{clusters}$
    \EndIf
\EndFor
\For{each cluster in $\mathrm{clusters}$}
    \State Compute fused box coordinates as a weighted average (weighted by confidence scores)
    \State Compute the final confidence score as the mean of the scores in the cluster
\EndFor
\State \Return final list of fused boxes
\end{algorithmic}
\end{algorithm}
\clearpage

\section{Host Relay (Algorithm Description)}\label{app:host-relay}
The host relay operates two concurrent reader threads, one per board, sharing a protected detection store. Algorithm~\ref{alg:relay-init} describes the top-level initialisation; Algorithms~\ref{alg:relay-readboard}--\ref{alg:relay-print-fused} describe the sub-procedures.

\begin{algorithm}[H]
\caption{Relay Initialisation}
\label{alg:relay-init}
\small
\begin{algorithmic}[1]
\State Open serial port for Board~1 at 115{,}200 baud with a 0.5~s per-call read timeout
\State Open serial port for Board~2 at 115{,}200 baud with a 0.5~s per-call read timeout
\State Initialise shared detection store $D \gets \emptyset$ and timestamp map $T \gets \emptyset$, both protected by a mutex
\State Spawn reader thread for Board~1, executing \Call{ReadBoard}{$\mathit{port}_1$}
\State Spawn reader thread for Board~2, executing \Call{ReadBoard}{$\mathit{port}_2$}
\State Block on thread join; on keyboard interrupt, close both ports and exit
\end{algorithmic}
\end{algorithm}

\begin{algorithm}[H]
\caption{ReadBoard(port)}
\label{alg:relay-readboard}
\small
\begin{algorithmic}[1]
\While{true}
    \State Read one byte $b$ from $\mathit{port}$
    \If{$b \neq \texttt{0xFE}$}
        \State Write $b$ to stdout as debug text
        \State \textbf{continue}
    \EndIf
    \State Read one byte $t$ from $\mathit{port}$
    \If{$t = \texttt{0xED}$}
        \State \Call{HandleDetection}{$\mathit{port}$}
    \ElsIf{$t = \texttt{0xEE}$}
        \State \Call{HandleSnapshot}{$\mathit{port}$}
    \Else
        \State Write bytes $b, t$ to stdout as debug text
    \EndIf
\EndWhile
\end{algorithmic}
\end{algorithm}

\begin{algorithm}[H]
\caption{HandleDetection(port)}
\label{alg:relay-handle-detection}
\small
\begin{algorithmic}[1]
\State Read 2 bytes: $\mathit{board\_id}, \mathit{num\_dets}$
\State Read $\mathit{num\_dets} \times 10$ bytes as raw detection payload
\State Initialise detection list $L \gets \emptyset$
\For{$i \gets 0$ \textbf{to} $\mathit{num\_dets} - 1$}
    \State Unpack five \texttt{uint16} values: $\tilde{x}_1, \tilde{y}_1, \tilde{x}_2, \tilde{y}_2, \tilde{s}$
    \State Decode $x_1 \gets \tilde{x}_1 / 4$, $y_1 \gets \tilde{y}_1 / 4$, $x_2 \gets \tilde{x}_2 / 4$, $y_2 \gets \tilde{y}_2 / 4$, and $\mathrm{score} \gets \tilde{s} / 10{,}000$
    \State Append $(x_1, y_1, x_2, y_2, \mathrm{score})$ to $L$
\EndFor
\State Acquire mutex; set $D[\mathit{board\_id}] \gets L$ and $T[\mathit{board\_id}] \gets \mathrm{now}$; release mutex
\State Print per-board detection summary to stdout
\State Acquire mutex; test whether both boards have timestamps within the last 10~s; release mutex
\If{both boards are fresh}
    \State \Call{PrintFused}{}
\EndIf
\end{algorithmic}
\end{algorithm}

\begin{algorithm}[H]
\caption{HandleSnapshot(port)}
\label{alg:relay-handle-snapshot}
\small
\begin{algorithmic}[1]
\State Read 4 bytes: $\mathit{board\_id}, W, H, \mathit{num\_dets}$
\State Read $\mathit{num\_dets} \times 10$ bytes using \Call{ReadExactly}{$\mathit{port}, \mathit{num\_dets} \times 10, 0.5~\mathrm{s}$}
\State Decode detections as in Algorithm~\ref{alg:relay-handle-detection}
\State Read $W \times H \times 3$ bytes using \Call{ReadExactly}{$\mathit{port}, W \times H \times 3, 30~\mathrm{s}$}
\If{fewer than $W \times H \times 3$ bytes were received}
    \State Discard frame and \Return
\EndIf
\State Reconstruct an RGB image of size $W \times H$ from the pixel data
\For{each detection $(x_1, y_1, x_2, y_2, \mathrm{score})$}
    \State Scale to snapshot space: $x'_1 \gets x_1(W/160)$, $y'_1 \gets y_1(H/160)$, $x'_2 \gets x_2(W/160)$, $y'_2 \gets y_2(H/160)$
    \State Draw the bounding box rectangle and score label on the image
\EndFor
\State Save annotated image to \texttt{captures/board<board\_id>\_<frame\_index>.png}
\end{algorithmic}
\end{algorithm}

\begin{algorithm}[H]
\caption{ReadExactly(port, n, timeout)}
\label{alg:relay-read-exactly}
\small
\begin{algorithmic}[1]
\State Initialise buffer $B \gets \emptyset$ and deadline $\gets \mathrm{now} + \mathit{timeout}$
\While{$\lvert B \rvert < n$}
    \If{$\mathrm{now} > \mathrm{deadline}$}
        \State \textbf{break}
    \EndIf
    \State Read up to $n - \lvert B \rvert$ bytes from $\mathit{port}$ into chunk $C$
    \State Append $C$ to $B$
\EndWhile
\State \Return $B$
\end{algorithmic}
\end{algorithm}

\paragraph{Rationale.}
At 115{,}200 baud, transferring a 19{,}200-byte camera frame takes approximately 1.67~s. Since \texttt{pyserial}'s \texttt{read(n)} returns after the per-call timeout (0.5~s) regardless of how many bytes have arrived, a naive single-call read recovers only approximately 5{,}760 bytes before returning. \textsc{ReadExactly} loops until the full payload is available.

\begin{algorithm}[H]
\caption{PrintFused()}
\label{alg:relay-print-fused}
\small
\begin{algorithmic}[1]
\State Acquire mutex; concatenate all detection lists in $D$ into a single list $A$; release mutex
\State Run WBF on $A$ with IoU threshold 0.45 and score threshold 0.40 to obtain fused list $F$
\State Print per-board detection counts and $\lvert F \rvert$ fused detections
\For{each $(x_1, y_1, x_2, y_2, \mathrm{score}) \in F$}
    \State Print bounding box coordinates and score
\EndFor
\end{algorithmic}
\end{algorithm}

The WBF procedure itself is described in Algorithm~\ref{alg:wbf} in Appendix~\ref{app:wbf}.
\clearpage

\section{Wi-Fi Peer-to-Peer (Algorithm Description)}\label{app:wifi-peer}
Each board runs an identical firmware image and differs only by the compile-time constant \texttt{BOARD\_ID}. Algorithm~\ref{alg:wifi-init} describes the one-time initialisation; Algorithm~\ref{alg:wifi-loop} describes a single iteration of the inference loop; Algorithms~\ref{alg:wifi-send-dets} and~\ref{alg:wifi-recv-dets} describe the UDP send and receive procedures. Relative to Appendix~\ref{app:host-relay}, the key architectural difference is that peer packets are self-filtered in firmware by checking \texttt{board\_id}, because no host relay is present to demultiplex the streams. The detection encoding is identical to the USB serial path described in Section~\ref{subsec:hardware-firmware}, and the WBF procedure itself is unchanged from Algorithm~\ref{alg:wbf} in Appendix~\ref{app:wbf}.

\begin{algorithm}[H]
\caption{Board Initialisation}
\label{alg:wifi-init}
\small
\begin{algorithmic}[1]
\Require $\mathit{BOARD\_ID} \in \{1, 2\}$, \texttt{WIFI\_SSID}, \texttt{WIFI\_PASSWORD}, model file \texttt{/models/yolo.tflite}
\State Load model bytes from the LittleFS filesystem into $\mathit{model\_data}$
\State Construct \texttt{MicroInterpreter} with the model, op resolver, and a 411~KB tensor arena
\State Call \texttt{AllocateTensors()} and verify INT8 input shape $[1, 160, 160, 3]$
\State Initialise the Himax camera at $160 \times 160$ RGB with $270^\circ$ rotation
\State Call \texttt{WiFi.begin(WIFI\_SSID, WIFI\_PASSWORD)}
\If{Wi-Fi initialisation succeeds}
    \State Create $\mathit{udp\_sock} \gets \texttt{socket(AF\_INET, SOCK\_DGRAM, 0)}$
    \State Bind $\mathit{udp\_sock}$ to \texttt{INADDR\_ANY}:\texttt{UDP\_PORT}
    \State Set socket option \texttt{SO\_RCVTIMEO} $\gets \texttt{UDP\_RX\_TIMEOUT\_MS}$ (8~ms)
    \State Set socket option \texttt{SO\_BROADCAST} $\gets 1$
    \State Set $\mathit{peer\_addr} \gets$ \texttt{255.255.255.255}:\texttt{UDP\_PORT}
    \State $\mathit{wifi\_ok} \gets \mathrm{True}$
\Else
    \State $\mathit{wifi\_ok} \gets \mathrm{False}$
\EndIf
\State $\mathit{ready} \gets \mathrm{True}$
\end{algorithmic}
\end{algorithm}

\begin{algorithm}[H]
\caption{Inference Loop (one iteration)}
\label{alg:wifi-loop}
\small
\begin{algorithmic}[1]
\Require $\mathit{ready} = \mathrm{True}$ and camera initialisation succeeded
\State Capture frame to $\mathit{nhwc}[160 \times 160 \times 3]$ via \texttt{Camera.grab()}
\For{each pixel $p$ and channel $c$ in $\mathit{nhwc}$}
    \State $f \gets ((p / 255) - \mu_c) / \sigma_c$ using ImageNet mean and standard deviation
    \State $q \gets \mathrm{clip}(\mathrm{round}(f / s_{\mathrm{in}}) + z_{\mathrm{in}}, -128, 127)$
    \State Write $q$ to the INT8 input tensor
\EndFor
\State Call \texttt{interpreter.Invoke()} and record $\mathit{invoke\_us}$
\State $\mathit{own\_dets}[0{:}n_{\mathrm{own}}] \gets \Call{DecodeDets}{\mathit{output\_tensor}}$ using sigmoid objectness, exponential box dimensions, and a score threshold of 0.40
\State $n_{\mathrm{peer}} \gets 0$
\If{$\mathit{wifi\_ok}$}
    \State \Call{SendDets}{$\mathit{own\_dets}, n_{\mathrm{own}}$} \Comment{Algorithm~\ref{alg:wifi-send-dets}}
    \State $n_{\mathrm{peer}} \gets \Call{RecvDets}{\mathit{peer\_dets}, \mathit{max\_n}}$ \Comment{Algorithm~\ref{alg:wifi-recv-dets}}
\EndIf
\State $\mathit{all\_dets} \gets \mathit{own\_dets} \cup \mathit{peer\_dets}[0{:}n_{\mathrm{peer}}]$
\State $n_{\mathrm{fused}} \gets \Call{WBF}{\mathit{all\_dets}}$ using Algorithm~\ref{alg:wbf} with IoU threshold 0.45 and score threshold 0.40
\State Set the status LED high if $n_{\mathrm{fused}} > 0$ or $n_{\mathrm{own}} > 0$
\State Print per-frame metrics to serial: own, peer, fused, \texttt{invoke\_us}, \texttt{wbf\_us}, \texttt{tx\_us}, \texttt{rx\_us}, arena KB
\end{algorithmic}
\end{algorithm}

\begin{algorithm}[H]
\caption{SendDets(dets, n)}
\label{alg:wifi-send-dets}
\small
\begin{algorithmic}[1]
\Require $\mathit{udp\_sock} \ge 0$ and $n \ge 0$
\State Allocate packet buffer of length $2 + 10n$ bytes
\State Write \texttt{WireHdr}: $\mathit{board\_id} \gets \mathit{BOARD\_ID}$ and $\mathit{num\_dets} \gets n$
\For{$i \gets 0$ \textbf{to} $n - 1$}
    \State Encode detection $i$ using the same fixed-point \texttt{WireDet} format as the USB serial path: $\hat{x}_1 \gets \lfloor 4x_1 \rfloor$, $\hat{y}_1 \gets \lfloor 4y_1 \rfloor$, $\hat{x}_2 \gets \lfloor 4x_2 \rfloor$, $\hat{y}_2 \gets \lfloor 4y_2 \rfloor$, and $\hat{s} \gets \lfloor 10000 s \rfloor$ as \texttt{uint16}
\EndFor
\State Call \texttt{sendto(udp\_sock, buf, pkt\_len, peer\_addr)} with destination \texttt{255.255.255.255}:\texttt{UDP\_PORT}
\State Record \texttt{udp\_tx\_us} and \texttt{udp\_tx\_bytes}
\end{algorithmic}
\end{algorithm}

\begin{algorithm}[H]
\caption{RecvDets(dets, max\_n)}
\label{alg:wifi-recv-dets}
\small
\begin{algorithmic}[1]
\Require $\mathit{udp\_sock} \ge 0$
\Ensure Returns the number of peer detections decoded, or 0 on timeout or self-packet
\State Call \texttt{recvfrom(udp\_sock, buf, ...)}; block for at most \texttt{SO\_RCVTIMEO} = 8~ms
\State Record \texttt{udp\_rx\_us}
\If{bytes received $\le \mathrm{sizeof}(\texttt{WireHdr})$}
    \State \Return 0
\EndIf
\State Read \texttt{WireHdr} from the buffer as $\mathit{board\_id}_{\mathrm{recv}}$ and $\mathit{num\_dets}$
\If{$\mathit{board\_id}_{\mathrm{recv}} = \mathit{BOARD\_ID}$}
    \State \Return 0
\EndIf
\State $n_d \gets \min(\mathit{num\_dets}, \mathit{max\_n})$
\For{$i \gets 0$ \textbf{to} $n_d - 1$}
    \State Decode \texttt{WireDet}[$i$]: $x_1 \gets \hat{x}_1 / 4$, $y_1 \gets \hat{y}_1 / 4$, $x_2 \gets \hat{x}_2 / 4$, $y_2 \gets \hat{y}_2 / 4$, and $s \gets \hat{s} / 10000$
    \State Write the decoded detection to $\mathit{dets}[i]$
\EndFor
\State Record \texttt{udp\_rx\_bytes}
\State \Return $n_d$
\end{algorithmic}
\end{algorithm}

\clearpage